\documentclass[lettersize,journal]{IEEEtran}
\usepackage{amsmath,amsfonts}
\usepackage{algorithmic}
\usepackage{algorithm}
\usepackage{array}
\usepackage[caption=false,font=normalsize,labelfont=sf,textfont=sf]{subfig}
\usepackage{textcomp}
\usepackage{stfloats}
\usepackage{url}
\usepackage{verbatim}
\usepackage{graphicx}
\usepackage{cite}
\usepackage{threeparttable}
\usepackage{multirow}
\usepackage{booktabs}
\usepackage{tabularx,array,wasysym}

\begin{document}

\title{The Evolution of Video Anomaly Detection: A Unified Framework from DNN to MLLM}

\author{
	\IEEEauthorblockN{
		Shibo Gao\IEEEauthorrefmark{1}\IEEEauthorrefmark{2},
		Peipei Yang\IEEEauthorrefmark{2}\IEEEauthorrefmark{3},
		Haiyang Guo\IEEEauthorrefmark{3},
		Yangyang Liu\IEEEauthorrefmark{2}\IEEEauthorrefmark{3},
		Yi Chen\IEEEauthorrefmark{2}\IEEEauthorrefmark{3}\IEEEauthorrefmark{4},
		Shuai Li\IEEEauthorrefmark{1}\IEEEauthorrefmark{2},
		Han Zhu\IEEEauthorrefmark{3},
		Jian Xu\IEEEauthorrefmark{2}\IEEEauthorrefmark{3},
		Xu-Yao Zhang\IEEEauthorrefmark{2}\IEEEauthorrefmark{3},
		Linlin Huang\IEEEauthorrefmark{1},
	} \\
	\IEEEauthorblockA{\IEEEauthorrefmark{1}Beijing Jiaotong University} \\
	\IEEEauthorblockA{\IEEEauthorrefmark{2}State Key Laboratory of Multimodal Artificial Intelligence Systems, Institute of Automation, Chinese Academy of Sciences} \\
	\IEEEauthorblockA{\IEEEauthorrefmark{3}School of Artificial Intelligence, University of Chinese Academy of Sciences}	\\
	\IEEEauthorblockA{\IEEEauthorrefmark{4} Zhongguancun Academy, Beijing, China}	
}

\maketitle

\begin{abstract}
Video anomaly detection (VAD) aims to identify and ground anomalous behaviors or events in videos, serving as a core technology in the fields of intelligent surveillance and public safety. With the advancement of deep learning, the continuous evolution of deep model architectures has driven innovation in VAD methodologies, significantly enhancing feature representation and scene adaptability, thereby improving algorithm generalization and expanding application boundaries. More importantly, the rapid development of multi-modal large language (MLLMs) and large language models (LLMs) has introduced new opportunities and challenges to the VAD field. Under the support of MLLMs and LLMs, VAD has undergone significant transformations in terms of data annotation, input modalities, model architectures, and task objectives. The surge in publications and the evolution of tasks have created an urgent need for systematic reviews of recent advancements. This paper presents the first comprehensive survey analyzing VAD methods based on MLLMs and LLMs, providing an in-depth discussion of the changes occurring in the VAD field in the era of large models and their underlying causes. Additionally, this paper proposes a unified framework that encompasses both deep neural network (DNN)-based and LLM-based VAD methods, offering a thorough analysis of the new VAD paradigms empowered by LLMs, constructing a classification system, and comparing their strengths and weaknesses. 
Building on this foundation, this paper focuses on current VAD methods based on MLLMs/LLMs. Finally, based on the trajectory of technological advancements and existing bottlenecks, this paper distills key challenges and outlines future research directions, offering guidance for the VAD community.
\end{abstract}

\begin{IEEEkeywords}
Video Anomaly Detection, Review, LLM, MLLM.
\end{IEEEkeywords}

\section{Introduction}
\IEEEPARstart{A}{nomalies} refer to events or entities that deviate from expectations and normal patterns, characterized by significant statistical deviations or semantic contradictions within specific spatiotemporal contexts.
Due to their exceptionally low occurrence frequency (in stark contrast to normal events), traditional anomaly detection techniques leverage machine learning to identify such rare patterns, enabling automated warning and recognition. VAD technology has permeated diverse fields, including financial fraud detection, network intrusion prevention, industrial defect screening, violent behavior recognition, medical lesion localization, and traffic violation monitoring~\cite{overall_1_ramachandra2020survey,overall_2_aggarwal2012applications,overall_3_jiang2023weakly}. 

Among these, video anomaly detection (VAD) focuses on identifying and grounding unconventional behaviors or events within video data. With its critical role in intelligent security and public administration, VAD has emerged as a cutting-edge topic of shared interest in both academia and industry~\cite{overall_4_pang2021deep,overall_5_yao2022dota}. VAD deconstructs anomalies into three dimensions of deviation: spatial (e.g., unauthorized object intrusion), temporal (e.g., anomalous movements such as reverse motion or sudden stops), and semantic (e.g., high-risk behaviors such as armed fighting). The continuous advancement of VAD technology has driven the intelligent upgrading of cross-domain security systems.

\begin{figure}[h]
	\centering
	\includegraphics[width=\linewidth]{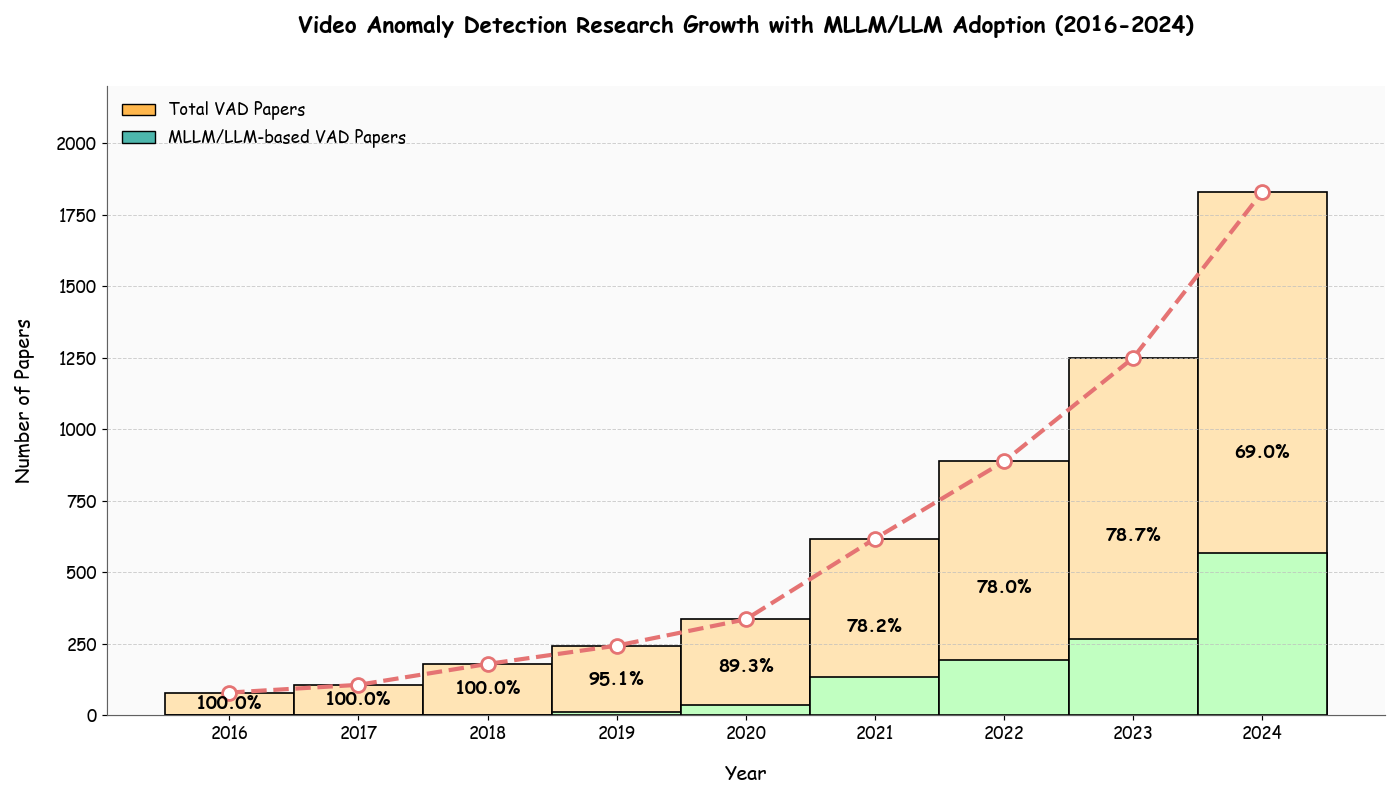}
	\caption{Annual distribution of VAD publications by methodology type. The result reveals a significant and continuous increase in the proportion of MLLM/LLM-based research, while traditional ML and DNN approaches have gradually declined, highlighting a paradigm shift toward multi-modal and large language model-driven techniques in the field.}
	\label{fig:google}
\end{figure}

Before the rise of deep learning, the field of VAD primarily relied on feature engineering-based solutions. Researchers typically hand-crafted discriminative features tailored to specific datasets.
These high-dimensional features were then processed using dimension reduction techniques, such as PCA or LDA, based on domain knowledge, followed by the construction of classifiers to distinguish between normal and anomalous samples~\cite{machine_1_cong2011sparse}. The main limitation of such methods lies in the lack of robustness of hand-crafted features, which constrained their detection performance in complex dynamic scenes. However, it is worth noting that these early studies provided significant paradigmatic references for subsequent deep learning-based detection methods. In particular, their exploration of feature abstraction and decision boundary construction holds valuable academic significance as a bridge between machine learning methods and recent methods.

With the rapid advancement of computer hardware performance and the breakthrough progress in deep learning technologies, the field of VAD has witnessed the emergence of various innovative deep learning-based methods.
Typical methods, such as ConvAE~\cite{convae_hasan2016learning}, learn representations of normal events by reconstructing video segments, while Future Predict~\cite{futurepredict_liu2018future} models normal patterns by predicting future frames, both leveraging image errors to identify anomalies. For weakly supervised scenarios with video-level labels, DeepMIL~\cite{deepmil_sultani2018real} introduces a novel multiple instance learning framework. Deep neural networks (DNNs), with their powerful feature representation and generalization capabilities, offer new solutions to address the severe imbalance between anomalous and normal samples. This enables precise anomaly detection across different supervision paradigms even under constrained annotation costs. Mainstream DNN-based VAD methods can be categorized into fully supervised VAD, unsupervised VAD, semi-supervised VAD, 
weakly supervised VAD, and Open-set VAD.

In recent years, the rapid development of multi-modal large language (MLLMs) and large language models (LLMs) has introduced transformative breakthroughs to the technical framework of VAD. Through multi-modal semantic understanding and generation capabilities, MLLMs and LLMs comprehensively upgrade the VAD framework across four dimensions: data annotation paradigms, modality fusion mechanisms, model architecture, and task objective.
At the \textbf{data annotation} level, traditional VAD relies on manually annotated pixel-level or frame-level labels. In contrast, MLLMs enable "text-video" supervised learning through cross-modal alignment, directly mapping natural language descriptions to anomaly semantic prototypes, significantly improving detection efficiency~\cite{cuva_du2024uncovering}.
At the \textbf{input modality} level, VAD evolves from single video stream analysis to multi-modal collaborative analysis, enhancing semantic understanding in complex scenarios~\cite{hawk_tang2024hawk}.
In terms of \textbf{model architecture}, MLLM/LLM-driven multi-modal pretraining frameworks enhance spatiotemporal-semantic joint anomaly detection capabilities through unified video-text representation spaces~\cite{vera_ye2025vera}.
At the \textbf{task objective} level, VAD extends from simple "detection-localization" to include "explainable diagnostics," "cross-modal retrieval," and "incremental learning," enabling more intelligent and responsive anomaly management mechanisms~\cite{holmes-vau_zhang2025holmes}.

\begin{figure}[h]
	\centering
	\includegraphics[width=\linewidth]{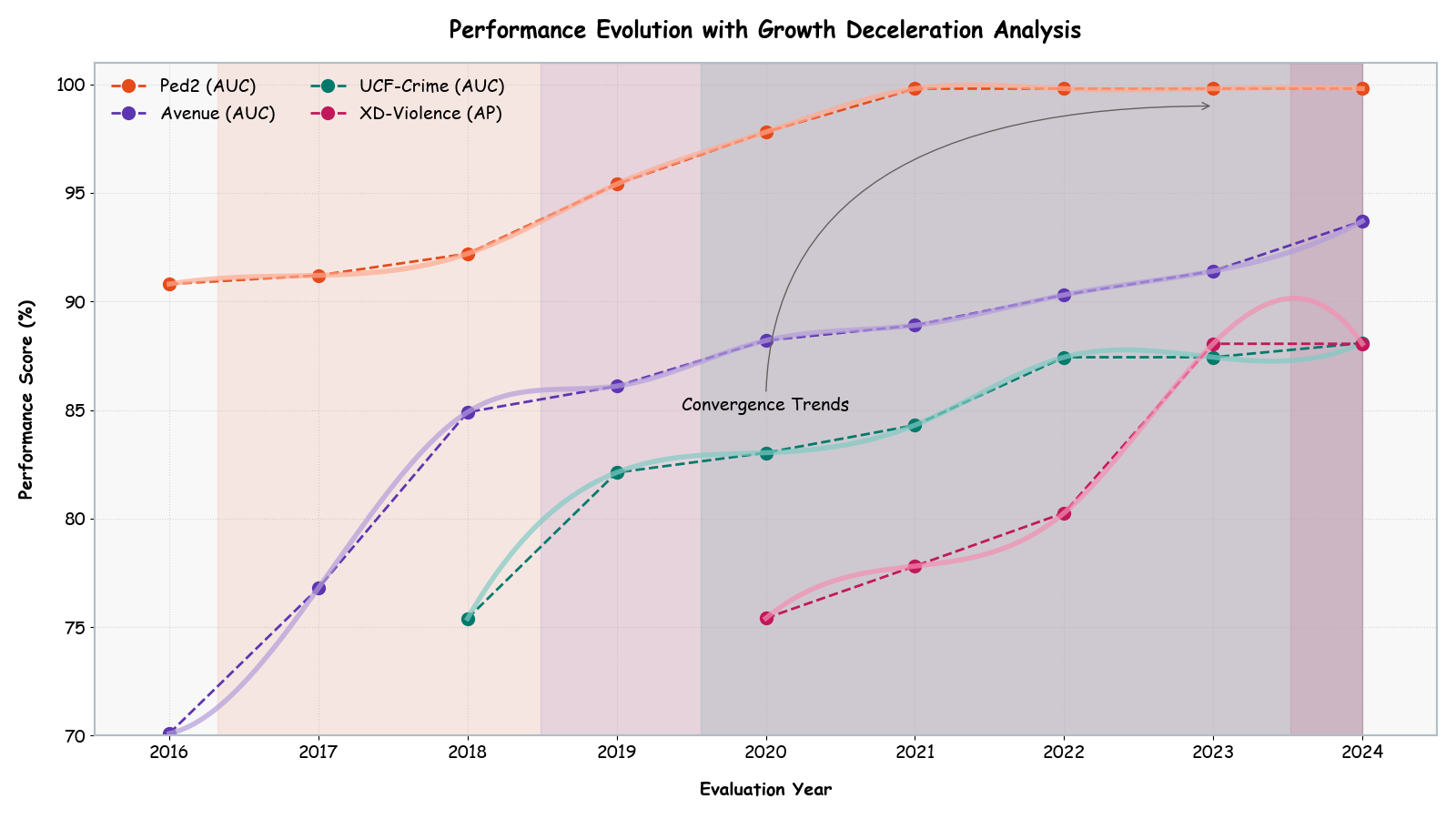}
	\caption{Performance evolution of representative DNN-based VAD methods on benchmark datasets. The results demonstrate that semi-supervised methods have reached near-saturation on simpler datasets, and weakly supervised methods show slowing progress on more complex datasets, indicating a performance bottleneck for conventional DNN approaches and underscoring the urgent need for new breakthroughs such as MLLM/LLM-driven VAD.}
	\label{fig:performance}
\end{figure}

%

To quantify the evolution trends in analysis techniques, we collected and analyzed the distribution of papers on video anomaly detection~(or related fields) over the past nine years from Google Scholar, as shown in Fig.~\ref{fig:google}. It is evident that the proportion of methods based on MLLM/LLM has been steadily increasing, while the share of traditional machine learning and DNN-based methods has been declining year by year.
Fig.~\ref{fig:performance} further illustrates the performance trends of two typical DNN paradigms on mainstream datasets. The AUC scores of semi-supervised VAD methods on Avenue~\cite{avenue} and Ped2~\cite{ped} datasets have reached saturation, while the performance breakthroughs of weakly supervised methods on complex datasets such as UCF-Crime~\cite{ucf-crime} and XD-Violence~\cite{xd-violence} have also slowed significantly.
This not only confirms the performance bottleneck of traditional DNN methods but also highlights the growing academic interest in MLLM/LLM-driven next-generation VAD technologies.

The aforementioned statistical data clearly demonstrate that MLLM/LLM-driven VAD has become a prominent research hotspot. In light of this, there is an urgent need to systematically categorize and comprehensively summarize the existing body of research. Such efforts would not only provide clear guidance for newcomers entering the field but also offer valuable references for experienced researchers.


\begin{table*}
	\caption{Comparison of recent VAD survey papers in terms of technical focus, field analysis, and task objectives. }
	\resizebox{\linewidth}{!}{
		\begin{tabular}{cccccccc}
			\toprule
			\multirow{4}{*}{Survey} & \multirow{4}{*}{Year}  & \multirow{4}{*}{Focus} & \multicolumn{3}{c}{Field Analysis}  & \multicolumn{2}{c}{Task Objective} \\
			\cmidrule(r){4-6} \cmidrule(r){7-8}
			& & &MLLM/LLM &Transformation &Framework &\multirow{2}{*}{VTG} &\multirow{2}{*}{VAU}\\
			& & &Based Methods & Analysis & Analysis & &\\
			\midrule
			Ramachandra et al.~\cite{survey_ramachandra2020survey} & 2020 & Semi-supervised VAD on single scene &\Circle &\Circle &\Circle &\CIRCLE &\Circle \\
			Nayak et al.~\cite{survey_nayak2021comprehensive} & 2021 & Semi-supervised VAD based on Deep Learning &\Circle &\Circle &\Circle &\CIRCLE &\Circle \\
			Tran et al.~\cite{survey_tran2022anomaly} & 2022 & Semi \& weakly supervised methods on Image \& Video &\Circle &\Circle &\Circle &\CIRCLE &\Circle \\
			Liu et al.~\cite{survey_liu2024generalized} & 2023 & Semi \& weakly supervised VAD based on DNN &\LEFTcircle &\Circle &\Circle &\CIRCLE &\Circle \\
			Wu et al.~\cite{survey_wu2024deep} & 2024 & Semi \& weakly \& un \& full supervised and open-set VAD &\LEFTcircle &\Circle &\Circle &\CIRCLE &\Circle \\
			Abdalla et al.~\cite{survey_abdalla2024video} & 2024 & The outlook and review of past VAD methods &\LEFTcircle &\Circle &\Circle &\CIRCLE &\LEFTcircle \\
			Ding et al.~\cite{survey_ding2024quo} & 2024 & LLMs and VLMs in VAD &\LEFTcircle &\LEFTcircle &\Circle &\Circle &\LEFTcircle \\
			Liu et al.~\cite{survey_liu2025networking} & 2025 & Networking systems for VAD &\LEFTcircle &\Circle &\Circle &\CIRCLE &\LEFTcircle \\
			\midrule
			Ours & 2025 & MLLM/LLM driven VAD and compatible framework analysis &\CIRCLE &\CIRCLE &\CIRCLE &\CIRCLE &\CIRCLE \\
			
			\bottomrule
		\end{tabular}
	}
	\label{table:survey}
\end{table*}

Based on this, we first collected and organized several representative VAD survey papers published in recent years, as summarized in Table.~\ref{table:survey}.
Among these, the survey by Ramachandra et al.~\cite{survey_ramachandra2020survey} primarily focuses on semi-supervised VAD in scenarios without scene transitions but does not explore more complex scenarios. Nayak et al.~\cite{survey_nayak2021comprehensive} conducted a comprehensive investigation into deep learning-based semi-supervised VAD methods but failed to cover other types of VAD methods. Tran et al.~\cite{survey_tran2022anomaly} reviewed emerging weakly supervised VAD methods; however, their study extended beyond VAD to include image anomaly detection, resulting in a lack of systematic organization and focus on VAD tasks. Liu et al.~\cite{survey_liu2024generalized} proposed structured classification frameworks, primarily addressing semi-supervised and weakly supervised VAD tasks, yet they did not incorporate recent advancements in the field.
More recently, Wu et al.~\cite{survey_wu2024deep} systematically organized and analyzed unsupervised, semi-supervised, weakly supervised, and open-set VAD tasks, providing detailed insights. However, it is regrettable that their review of MLLM/LLM-based methods was not in-depth, offering only a brief overview and overlooking the transformative impact of MLLM/LLM technologies on the VAD domain.

To systematically address the existing gaps in the current research landscape, we have conducted a comprehensive and in-depth investigation into VAD in the context of the MLM/LLM era. This study focuses on several key dimensions, aiming to provide a thorough and detailed analysis of MLLM/LLM-empowered VAD research.

\textbf{First}, we deeply analyzed the core value and profound impact of semantic information embedded in MLLMs and LLMs on the VAD domain, elucidating the intrinsic driving mechanisms behind the transformative changes in VAD technologies. Leveraging their powerful semantic understanding and generation capabilities, MLLMs and LLMs have injected new vitality into VAD tasks. We meticulously deconstructed this process, revealing how semantic information enhances the performance of VAD systems in accomplishing their objectives.
\textbf{Second}, we revisited and redefined the VAD task, proposing a more comprehensive and adaptable unified analytical framework. This framework is compatible with both traditional DNN-based methods and cutting-edge methods based on MLLMs and LLMs. Using this framework, we quantitatively compared different technological pathways in terms of performance, adaptability to application scenarios, and computational resource consumption.
\textbf{Third}, following the proposed unified analytical framework, we focused on introducing MLLM/LLM-based VAD methods, aiming to provide researchers in the field with a clear and practical research roadmap, helping them avoid unnecessary detours and significantly improve research efficiency.
\textbf{Finally}, based on a thorough review of the evolution of VAD technologies and precise insights into current bottlenecks, we distilled the key challenges facing the VAD domain and offered forward-looking perspectives on potential breakthrough directions. These projections provide clear research guidance for the VAD research community, helping researchers identify promising directions and focus their efforts effectively in future work.

Given that previous surveys have already conducted exhaustive investigations into traditional DNN-based VAD methods, and considering the relatively limited recent progress in this area, we have concentrated this study on VAD research in the MLLM/LLM era, which can provide robust academic support for this emerging and highly promising research direction.

The main contributions of this survey can be summarized as follows:
\begin{itemize}
	\item We conducted a detailed analysis of the transformations in the VAD field driven by advancements in MLLMs and LLMs. From a fundamental logical perspective, we explained how semantic information acts as a driving force behind the technological paradigm shift in VAD.
	\item We revisited and redefined the VAD task, introducing a more comprehensive and adaptable unified analytical framework. Building on this foundation, we provided an in-depth introduction to MLLM/LLM-based VAD methods. To the best of our knowledge, this is the first comprehensive survey dedicated to MLLM/LLM-based approaches for VAD.
	\item Based on the current state of research, we distilled the key challenges facing the VAD field and provided forward-looking insights into potential breakthrough directions.
\end{itemize}

\section{Analysis of Video Anomaly Detection}
\label{sec:analysis}

In the early stages of research on VAD methods based on deep neural networks, researchers typically extracted modality features of normal and anomalous events and compared the test set to these features to detect anomalies. These methods predominantly operated in the visual feature space to identify anomalies~\cite{futurepredict_liu2018future,astrid2021learning,wang2022making}. For instance, in semi-supervised VAD, reconstruction or prediction paradigms were utilized to design self-supervised tasks, enabling network models to learn the features of normal video frames~\cite{park2020learning,hu2022pedestrian}. Similarly, in weakly supervised VAD, the one-stage multiple instance learning (MIL) paradigm was employed to identify the most likely anomalous and normal segments, thereby learning classification boundaries within the visual feature space~\cite{156_zhu2019motion,157_zhang2019temporal,158_liu2022collaborative}.

\begin{figure}[h]
	\centering
	\includegraphics[width=\linewidth]{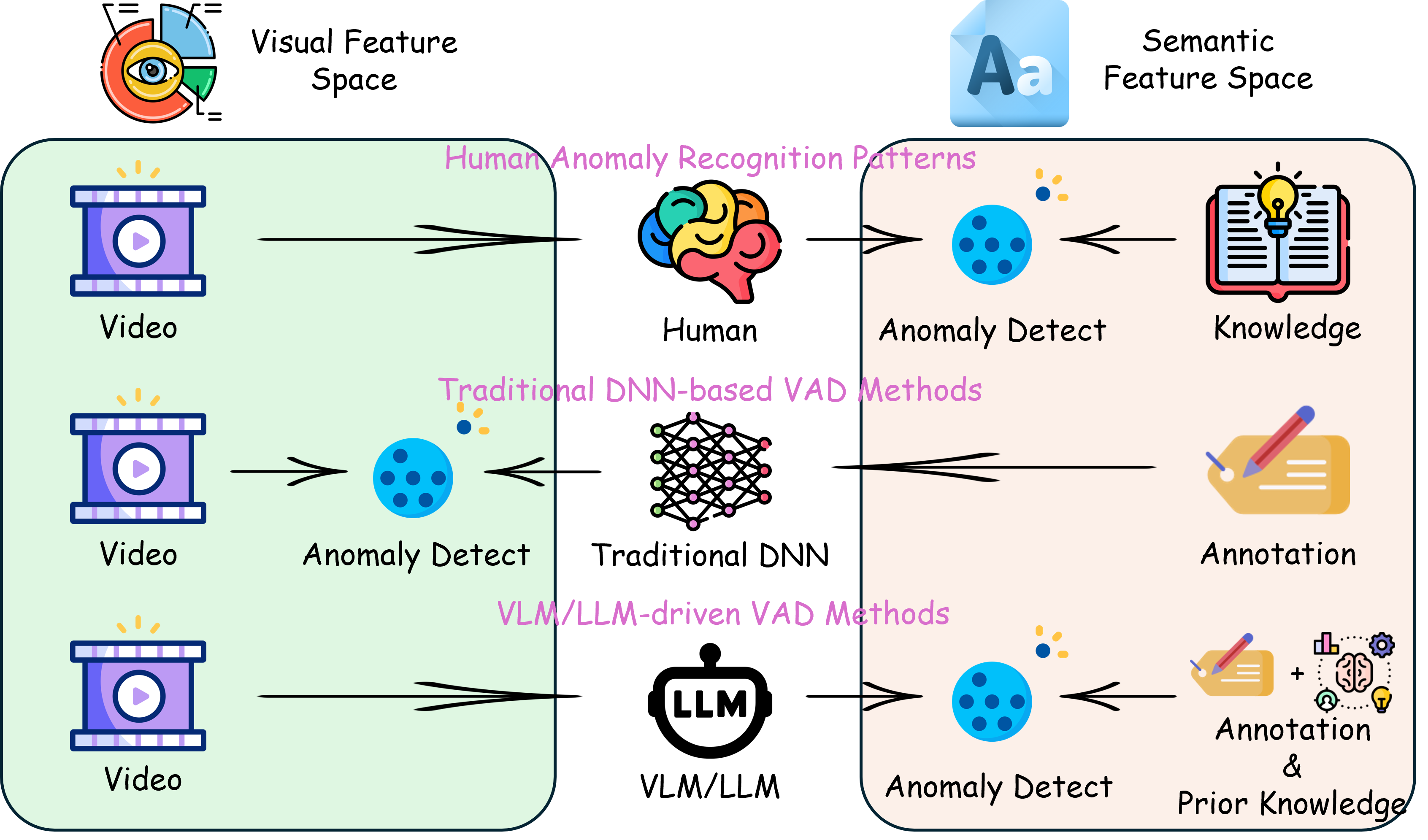}
	\caption{Comparison of different approaches for anomaly detection:
		(1) Human brain’s semantic-space reasoning: Visual information is abstracted into semantic features, and anomalies are detected by rule-based reasoning within the semantic space;
		(2) Traditional DNN paradigm: Annotation information is mapped to the visual feature space to learn classification boundaries, but this approach is limited in generalization and interpretability;
		(3) VLM/LLM paradigm: Leveraging rich pretrained semantic knowledge and prompt mechanisms, classification boundaries are directly constructed in the semantic space, enabling superior generalization and interpretability.}
	\label{fig:VandS}
\end{figure}

We decompose the VAD task into three components: video streams, data annotations, and network architecture. Video streams and data annotations provide different levels of visual and semantic information, respectively, while the network architecture bridges the visual feature space and the semantic feature space, learning classification boundaries in one of these spaces. Traditional DNN-based methods typically map annotations from the semantic feature space to the visual feature space and then learn classification boundaries in the latter. However, this mapping process has two fundamental limitations: first, the manually predefined mapping rules are influenced by the biases of the annotation information, which restricts the generalization ability of the model; second, the implicit use of semantic information significantly limits the interpretability of these models. The second row of Fig.~\ref{fig:VandS} illustrates this process.

At this stage of research, the high labeling costs, poor generalization capabilities, and lack of interpretability have become the main bottlenecks constraining the development of VAD. From a real-world perspective, the severe imbalance between anomalous and normal samples makes high labeling costs almost unavoidable. Moreover, learning classification boundaries in the visual feature space often leads to a lack of interpretability for anomalies, further limiting the generalization capacity of models. As a result, research at this stage has primarily focused on improving model performance in specific datasets or limited scenarios.

In contrast, the way the human brain processes anomalies is fundamentally different from the aforementioned methods. Upon receiving visual information, humans first abstract it into semantic representations. Then, depending on the task requirements, humans compare the abstracted semantic information with predefined rules in the semantic space to identify anomalies. In this process, the predefined rules serve as semantic annotations, while the human brain functions as the network architecture. This anomaly detection method, based on the semantic space, offers a high degree of interpretability. Moreover, leveraging the brain's extensive prior knowledge allows for the construction of classification boundaries in the semantic space that are more reasonable and generalizable. The first row of Fig.~\ref{fig:VandS} illustrates this process.

With the rapid advancement of VLMs and LLMs, VAD research has increasingly focused on leveraging these models. Early works employing VLMs utilized the prior knowledge learned during model pretraining to directly detect anomalies in the semantic space. For example, VadCLIP~\cite{vadclip} employs a pretrained CLIP model and learnable prompts to achieve anomaly detection and classification in the semantic space. Compared to traditional DNN-based VAD methods, these approaches not only demonstrate superior performance but also exhibit stronger generalization capabilities. Although these early methods were limited by the constraints of the pretrained models and lacked strong interpretability, they laid a solid foundation for subsequent breakthroughs in the VAD field.

We argue that three key characteristics of LLMs and MllMs are particularly significant for the VAD domain. First, LLMs/MLLMs can directly abstract visual signals into comprehensible semantic information~\cite{lavad,cuva_du2024uncovering,followtherules_yang2024follow}. Second, these models have already acquired rich semantic knowledge during the pretraining phase~\cite{llama,internvl}. Finally, they allow for seamless interaction through prompt-based approaches, eliminating the need for additional training overhead~\cite{suvad,vera_ye2025vera}. 

These three advantages drastically reduce the complexity of many previously challenging problems. The optimization of the discriminative space not only significantly enhances the interpretability of models but also enables the construction of more flexible classification boundaries. Furthermore, the rich semantic knowledge accumulated during pretraining, combined with the prompt-based interaction mechanism, endows these models with exceptional generalization capabilities. The third row of Figure~\ref{fig:VandS} visually illustrates this process.

Notably, some training-free methods based on VLMs/LLMs (e.g., LAVAD~\cite{lavad} and SUVAD~\cite{suvad}) achieve performance comparable to traditional approaches without requiring any additional training. Moreover, these methods provide additional functionalities, such as anomaly explanation, scene generalization, and anomaly adjustment.

Overall, the rich prior semantic knowledge learned by VLMs/LLMs during the pretraining phase, combined with their inherent interpretability, allows current VAD methods to construct superior classification boundaries in the semantic space or directly identify anomalies. Although the computational cost of these models is significantly higher than that of traditional DNN-based methods, their unique advantages align perfectly with the demands of real-world VAD applications. 
We will continue to analyze and reveal the changes brought by VLM/LLM to VAD in the subsequent detailed explanation of the methods.

\section{Background}
\label{sec:background}

In this section, we will first introduce the traditional objectives of the VAD task and discuss how these objectives have been expanded under the influence of MLLMs and LLMs. 

Following this, we will present a novel VAD classification framework that is compatible with both traditional DNN-based VAD methods and MLLM/LLM-driven methods. Building on this framework, we will further clarify the definitions of VAD tasks under different supervision settings. Finally, we will introduce the datasets and evaluation metrics commonly used in the VAD.

\subsection{Task Objective}
\label{task objective}
In general, the task objectives of Video Anomaly Detection can be broadly categorized into two main aspects: Video Anomaly Grounding (VTG) and Video Anomaly Understanding (VAU).

\begin{figure*}[h]
	\centering
	\includegraphics[width=\linewidth]{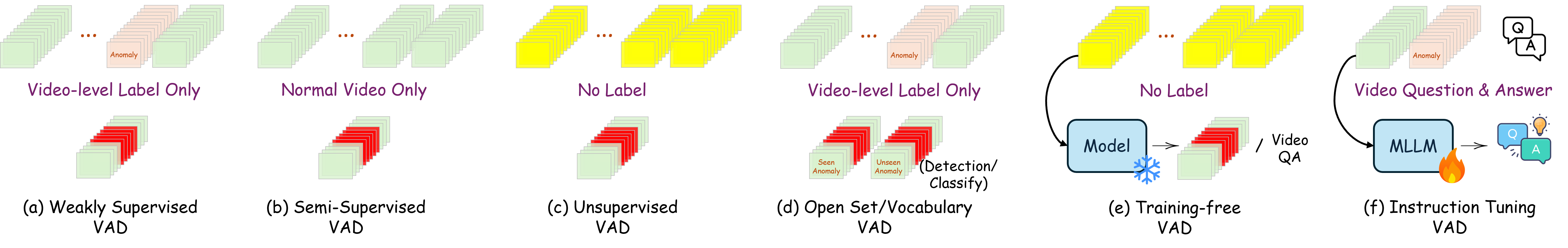}
	\caption{Illustration of the differences in training and testing setups among various VAD tasks under different supervision paradigms. Each supervision type is characterized by distinct requirements for input videos and label annotations during both training and inference.}
	\label{fig:difference}
\end{figure*}

\textbf{Video Anomaly Grounding.} The goal of the VTG task branch is to identify which video frames contain anomalies. Specifically, in the methods that require training, a series of training videos and their corresponding label $\mathcal{Y}_{train}$ (which vary depending on the level of supervision) are provided:
\begin{equation}
	\mathcal{X}_{train} = \{x_{train,t}\}_{t=1}^T,~ x_{train,t} \in \mathbb{R}^{H \times W \times C}.
\end{equation}
The detection model $\Phi(\theta)$ takes the training videos $\mathcal{X}_{train}$ as input and is optimized according to the following objective:
\begin{equation}
	\min\limits_{\theta} l = \mathcal{P}\{\Phi(\theta;\mathcal{X}_{train}),\mathcal{Y}_{train}\},
\end{equation}
where $\mathcal{P}$ represents a certain distance or divergence function used to quantify the discrepancy between the predictions and labels.
During the inference process, given the test videos $\mathcal{X}_{test}$ and its corresponding annotations $\mathcal{Y}_{test}$:
\begin{equation}
	\mathcal{X}_{test} = \{x_{test,t}\}_{t=1}^T,~ x_{test,t} \in \mathbb{R}^{H \times W \times C},
\end{equation}
\begin{equation}
	\mathcal{Y}_{test} = \{y_{test,t}\}_{t=1}^{T},~ y_{test,t} \in \{0,1\}.
\end{equation}
The model's performance is generally evaluated through the following process:
\begin{equation}
	\mathcal{S} = \mathcal{L}(\Phi(\mathcal{X}_{test}), \mathcal{Y}_{test}),
\end{equation}
where $\mathcal{L}$ measures the difference between the model's predictions and the ground truth.

\textbf{Video Anomaly Understanding.} In the VAU task branch, traditional DNN-based methods require the model to correctly classify anomalous frames under specific supervised conditions. With the advancement of MLLMs/LLMs, the VAU task has been further expanded to include anomaly description and causal understanding of anomalies.

The task process of VAU is similar to that of the aforementioned VTG task, with the primary difference being the label information provided during the testing phase:
\begin{equation}
	\mathcal{Y}_{test} \in \mathcal{N},
\end{equation}
where $\mathcal{N}$ represents the set of semantic anomaly annotations, including video-level category labels, anomaly descriptions, and causal analysis of anomalies.

Notably, VTG and VAU tasks are not independent of each other. For example, in weakly supervised VAD tasks, multi-task learning is often employed to simultaneously achieve anomaly grounding and anomaly classification. Additionally, training-free methods based on MLLMs/LLMs typically rely on the collaboration of multiple large models to accomplish both VTG and VAU tasks simultaneously.

\subsection{VAD Tasks under Different Supervision Methods}

Traditional DNN-based VAD methods can be categorized into five distinct types based on the supervision signal: semi-supervised VAD (SVAD), weakly supervised VAD (WVAD), fully supervised VAD (FVAD), unsupervised VAD (UVAD), and open-set supervised VAD (OSVAD). Due to the difficulty of collecting anomalous data and the labor-intensive nature of annotation, fully supervised VAD has gradually fallen out of favor. Meanwhile, the rapid development of VLMs and LLMs has introduced new categories to VAD, including open-vocabulary VAD (OVVAD), training-free VAD (TVAD), and instruction fine-tuning VAD (IFVAD).

Although the task objectives of VAD under different supervision paradigms are similar, they involve distinct training and testing setups. We illustrate these differences in Fig.~\ref{fig:difference}. Based on the task objectives introduced in Section.~\ref{task objective}, the seven types of VAD can be categorized as follows, according to the variations in input videos $\mathcal{X}$ and labels $\mathcal{Y}$:

\textbf{Semi-supervised VAD}. Semi-supervised VAD assumes that training videos only contain normal events during the training phase, i.e.,

\begin{equation}
	\mathcal{Y}_{train} = \{y_{train,t}\}_{t=1}^{T},~ y_{train,t} = 0.
\end{equation}
Under this supervision paradigm, the model learns the normal patterns from normal data and treats samples deviating from these patterns as anomalies. In this setting, datasets typically exhibit no scene transitions or only a minimal number of transitions. Due to the absence of anomalous samples during training, the annotation cost for semi-supervised VAD is relatively low. However, training exclusively on normal samples may lead the model to classify any samples not present in the training set as anomalies, resulting in a high false positive rate. Moreover, this paradigm is often tightly coupled with specific scenes, meaning that even minor deviations, such as slight camera rotations, can cause catastrophic performance degradation.

\textbf{Weakly supervised VAD}. Compared to semi-supervised VAD, weakly supervised VAD provides stronger supervision signals. During training, both normal and anomalous videos are provided, but the anomalous videos only contain video-level labels. In other words, the model will be not provided the exact timestamps of anomalous events in the videos, i.e.,

\begin{equation}
	\mathcal{Y}_{train,n} = \{y_{train,n,t}\}_{t=1}^{T},~ y_{train,t} = 0,
\end{equation}
\begin{equation}
	\mathcal{Y}_{train,a} = A,~A \in \mathcal{A},
\end{equation}
where $\mathcal{A}$ represent the set of anomaly classes contained in the dataset.
Due to the stronger supervision signals, weakly-supervised VAD usually achieves better performance than semi-supervised VAD and exhibits some adaptability to scene transitions. However, this paradigm imposes higher requirements on algorithm design, and collecting anomalous videos requires additional effort.

\textbf{Unsupervised VAD}. Unsupervised VAD aims to detect anomalies directly from completely unlabeled videos in an unsupervised manner. Under this paradigm, no division between training and testing sets is required, i.e.,

\begin{equation}
	\mathcal{X} = \mathcal{X}_{test},~\mathcal{Y}_{train} = \emptyset.
\end{equation}
Unsupervised VAD allows the model to continuously update itself during anomaly detection without relying on any data collection or annotation process. However, due to the lack of labels, unsupervised algorithms are often more complex and tend to exhibit inferior detection performance.

\textbf{Open-set VAD}. Open-set VAD aims to detect anomalies that are unseen during training. Specifically, the training set includes normal samples and anomalous samples with video-level labels,
\begin{equation}
	\mathcal{Y}_{train,n} = \{y_{train,n,t}\}_{t=1}^{T},~ y_{train,t} = 0,
\end{equation} 
\begin{equation}
	\mathcal{Y}_{train,a} = A,~A \in \mathcal{A}_{base},
\end{equation}
where $\mathcal{A}_{base}$ represents the basic seen category of anomalies.

The test set consists of both anomalies seen during training and unseen anomalies, referred to as "seen anomalies" and "unseen anomalies," respectively, i.e.,
\begin{equation}
	\mathcal{Y}_{train,a} = A,~A \in \mathcal{A},~ \mathcal{A} = \mathcal{A}_{base} \cup \mathcal{A}_{novel},
\end{equation}
where $\mathcal{A}_{novel}$ represents the unseen category of anomalies. It is worth noting that open-set VAD typically does not require identifying the exact category of unseen anomalies. Compared to mainstream semi-supervised and weakly-supervised VAD, open-set VAD demonstrates strong scene generalization and holds significant value for real-world applications. However, this superior performance often relies on specially designed additional modules or the construction of pseudo anomalies to detect unseen anomalies.

\textbf{Open-vocabulary VAD}. Open-vocabulary VAD aims to precisely classify anomalies that have not been encountered in the training set. Specifically, the training set provides normal samples and anomalous samples with detailed labels. The test set includes anomalies seen in the training set as well as those unseen, referred to as visible anomalies and unseen anomalies. These are consistent with open-set VAD.
Unlike open-set VAD, open-vocabulary VAD requires not only detecting unseen anomalies but also identifying their specific categories. This setting enhances the model's adaptability in diverse scenarios, making it particularly effective in handling new types of anomalies in real-world applications. However, achieving open vocabulary VAD typically requires sophisticated model design and heavily relies on the rich prior knowledge learned by pre-trained models.

\textbf{Training-free VAD}. Training-free VAD aims to leverage the powerful prior knowledge of VLMs or LLMs for anomaly detection. Specifically, training-free VAD analyzes videos directly based on preset rules or general knowledge without requiring any adjustment to model parameters, represented as
\begin{equation}
	\Phi(\theta;\mathcal{X}) = \Phi(\emptyset;\mathcal{X}).
\end{equation}
In this setting, the detection system can identify anomalous events without relying on specific training data, while ensuring a certain level of detection performance. The advantage of training-free VAD lies in its efficiency and flexibility, enabling rapid adaptation to new scenarios and new types of anomalies, thereby significantly enhancing practical applicability. However, despite the absence of additional training, the model's performance remains dependent on the quality of predefined rules and prior knowledge. Moreover, while the VLMs/LLMs employed in this method exhibit strong capabilities, they are associated with high computational costs.

\textbf{Instruction fine-tuning VAD}. Instruction fine-tuning VAD aims to fine-tune pre-trained large models to optimize their performance in VAD tasks. Specifically, instruction fine-tuning VAD involves training the model on specific datasets, adjusting its parameters to better adapt to particular scenarios and task requirements. During this process, the model learns the characteristics of anomalous samples and adjusts its behavior based on user-specific instructions, thereby improving detection accuracy and robustness. The advantage of instruction fine-tuning VAD lies in leveraging the knowledge of pre-trained models while achieving personalized and targeted anomaly detection, enhancing the model's effectiveness in practical applications. However, this approach requires additional training data and computational resources, and the fine-tuning process may reduce the model's generalization ability on new tasks. Moreover, excessive fine-tuning may lead to overfitting the model to specific datasets, thereby compromising its performance in diverse scenarios.

\begin{table*}
\caption{Comparison of mainstream VAD datasets in terms of domain, scale, annotation granularity, and support for advanced semantic or audio-based understanding.}
\label{table:dataset}
\resizebox{\linewidth}{!}{
\begin{tabular}{ccc|ccc|ccc}
	\toprule 		
	\multirow{2}{*}{Dataset} & \multirow{2}{*}{Domain} & \multirow{2}{*}{Year} & \multicolumn{3}{c|}{Dataset Statistical Information}  & \multicolumn{3}{c}{Dataset Annotation} \\
	\cline{4-9}
	& & & Video Samples & Total Frames & Anomaly Categories & Location & Understanding & Audio \\
	
	\midrule
	
	Subway Entrance~\cite{subway} &Streetscape & &1 &86,535 &5 &Frame & &  \\
	
	Subway Exit~\cite{subway} &Streetscape & &1 &38,940 &3 &Frame & &  \\
	
	UMN~\cite{umn} &Behaviors & &5 &3,855 &1 &Frame & & \\
	
	UCSD Ped1~\cite{ped} &Streetscape & &70 &14,000 &5 &Bounding-box & & \\
	
	UCSD Ped2~\cite{ped} &Streetscape & &28 &4,560 &5 &Bounding-box & & \\
	
	CUHK Avenue~\cite{avenue} &Streetscape & &37 &30,652 &5 &Bounding-box & & \\
	
	Street Scene~\cite{street} &Traffic & &81 &203,257 &17 &Bounding-box & & \\
	
	NWPU Campus~\cite{nwpu} &Streetscape & &547 &1,466,073 &28 &Frame & & \\
	
	ShanghaiTech~\cite{shanghaitech} &Streetscape & &437 &317,398 &13 &Bounding-box & & \\
	
	\midrule
	
	UCF-Crimes~\cite{ucf-crime} &Crime & &1900 &13,741,393 &13 &Frame &Classify & \\
	
	UCf-Crime Extension~\cite{ucf-crime-extension} &Crime & &2183 &14,475,793 &15 &Frame &Classify & \\
	
	XD-Violence~\cite{xd-violence} &Violence & &800 &114,096 &6 &Frame &Classify &\checkmark \\
	
	
	TAD~\cite{tad} &Traffic & &344 &721,280 &4 &Bounding-box & & \\
	
	BOSS~\cite{boss} &Multiple & &16 &48,624 &11 &Frame &Classify &\checkmark \\
	
	CamNuvem~\cite{camnuvem} &Robbery & &486 &6,151,788 &1 &Frame & & \\
	
	UCVL(Not released)t~\cite{ucvl} &Crime & &1699 & &13 &Frame &Classify & \\
	
	DoTA~\cite{dota} &Traffic & &4677 &731,932 &1 &Frame & & \\
	
	\midrule
	
	Ubnormal~\cite{ubnormal} &Multiple & &543 &236,902 &22 &Pixel &Classify & \\
	
	\midrule
	
	CUVA~\cite{cuva_du2024uncovering} &Multiple & &1000 &3,345,097 &11 &Time Duration &Video QA &\checkmark \\
	
	ECVA~\cite{ecva} &Multiple & &2500 &19,042,560 &21 &Time Duration &Video QA &\checkmark \\
	
	VANE-Bench~\cite{vane-bench} &Multiple & &325 &951,482 &19 & &Video QA & \\
	
	VAGU(Not released) &Multiple & &7567 & &21 &Time Duration &Video QA &\checkmark \\
	
	PreVAD(Not released)~\cite{prevad} &Multiple & &35279 & &35 &Frame &Video QA & \\
	
	HIVAU-70k~\cite{holmes-vau_zhang2025holmes} &Multiple & &5443 &13,855,489 &15 & &Video QA &\checkmark \\
	
	UCA~\cite{uca} &Crime & &1854 &11,817,597 &13 &Frame &Video QA &\checkmark \\
	
	\bottomrule

\end{tabular}
}
\end{table*}

\subsection{Datasets and Metrics}

For various VAD tasks, the existing literature provides a wealth of publicly available datasets that encompass diverse scenarios. Specifically, these datasets range from those focused on specific scenes to those covering a wide variety of complex situations. Moreover, differences exist across datasets in terms of the number of videos, average video duration, anomaly categories, and annotation methods. In Table.~\ref{table:dataset}, we provide a detailed comparison of the current mainstream datasets. It is worth noting that some datasets were initially designed for specific VAD tasks; however, with the continuous advancement of technology, these datasets can now also be applied to other VAD tasks. For instance, any semi-supervised or weakly supervised VAD dataset can be utilized for evaluating training-free VAD methods.

Corresponding to the two task objectives of VAD, the evaluation metrics for VAD tasks are also divided into two categories. One focuses on assessing the model's ability to ground anomalous events in the temporal dimension, while the other evaluates the model's ability to understand anomalous events in the semantic dimension.

In the temporal dimension, commonly used evaluation metrics include AUC~\cite{deepmil_sultani2018real}, ERR/EDR~\cite{avenue}, and Accuracy~\cite{accuracy}. Additionally, there are metrics designed for detecting anomalous regions and trajectories, such as RBDC and TBDC~\cite{street}.

In the semantic dimension, the most commonly used metric is AP~\cite{xd-violence}. With the growing influence of VLMs and LLMs, VAD has also adopted some metrics to evaluate the performance of large models, such as BLEU~\cite{bleu} and ROUGE~\cite{rouge}. Furthermore, considering the complexity of video understanding tasks, particularly anomalous video understanding, some scoring metrics based on LLMs or MLLMs have also been proposed~\cite{holmes-vau_zhang2025holmes,cuva_du2024uncovering,suvad}.

\textbf{AUC (Area Under the Curve)~\cite{deepmil_sultani2018real}}. AUC refers to the area under the ROC (Receiver Operating Characteristic) curve. The ROC curve is plotted by comparing the True Positive Rate (TPR) and the False Positive Rate (FPR) across different thresholds:
\begin{equation}
	TPR = \frac{TP}{TP + FN},~ FPR = \frac{FP}{FP + TN}.
\end{equation}

TP (True Positive) is the number of samples that are actually positive and correctly predicted as positive, TN (True Negative) is the number of samples that are actually negative and correctly predicted as negative, FP (False Positive) is the number of samples that are actually negative but incorrectly predicted as positive, and FN (False Negative) is the number of samples that are actually positive but incorrectly predicted as negative.

\textbf{EER (Equal Error Rate) and EDR (Equal Detected Rate)~\cite{avenue}}. EER refers to the error rate at the point where the FPR equals the FNR on the ROC curve. EDR represents the proportion of anomalies detected by the system under a specific detection threshold relative to the total number of anomalies. EER provides a balance point, while EDR emphasizes the completeness of detection. Particularly in anomaly detection, a high recall rate is crucial, as missing a true anomaly could have far more severe consequences than mistakenly labeling a normal event as anomalous.

\textbf{Accuracy~\cite{accuracy}}. Accuracy is a performance metric used in classification models or diagnostic tests, defined as the ratio of correct predictions to the total number of predictions:
\begin{equation}
	Accuracy = \frac{TP + TN}{TP + TN + FP + FN}.
\end{equation}
The definitions of TP, TN, FP, and FN remain consistent with those provided earlier. While accuracy is an intuitive metric for evaluating performance, relying solely on accuracy can be misleading in scenarios involving class imbalance. Consequently, within the domain of VAD, accuracy is less frequently employed compared to metrics such as the AUC.

\textbf{RBDC (Region-Based Detection Criterion)~\cite{street}}.
RBDC assesses the capability of a model to precisely localize anomalous regions within individual video frames. This metric computes scores by comparing the detected anomalous regions with annotated ground-truth regions:
\begin{equation}
	RBDC = \frac{Region_{model} \cap Region_{gt}}{Region_{gt}},
\end{equation}
where $Region_{model}$ represents the anomalous areas detected by the model, while B$Region_{gt}$ represents the ground truth.
A higher RBDC score indicates superior spatial localization performance of the model, meaning it can more accurately pinpoint the locations of anomalies within video frames.

\textbf{TBDC (Track-Based Detection Criterion)~\cite{street}},
TBDC evaluates the model's capability for detecting and tracking anomalies along the temporal dimension, measuring performance in grounding anomalies across consecutive video frames:
\begin{equation}
	TBDC = \frac{Track_{model} \cap Track_{gt}}{Track_{gt}},
\end{equation}
where $Track_{model}$ represents the anomalous tracks detected by the model, while B$Track_{gt}$ represents the ground truth. 
This metric is particularly suited for scenarios where anomalous events exhibit temporal continuity, such as objects moving anomalously or events spanning multiple frames.
TBDC employs Intersection-over-Union (IoU) to quantify the overlap between predicted trajectories and ground-truth trajectories and emphasizes temporal continuity, thus ensuring that the model is capable not only of detecting anomalies in individual frames but also accurately tracking anomalies throughout the entire video sequence.

\textbf{AP (Average Precision)~\cite{xd-violence}}. AP refers to the area under the precision-recall curve. Precision represents the proportion of correctly identified positive samples, while recall (or sensitivity) measures the proportion of positive samples that are correctly identified. AP is particularly effective in scenarios with a limited number of positive samples (e.g., anomalous samples). In the VAD, this metric not only focuses on anomaly classification but also emphasizes the ability to localize anomalies at the video frame level. Nevertheless, for the sake of comparison with AUC, AP is categorized here as an evaluation metric for anomalous video understanding.

\textbf{BLEU~\cite{bleu}, ROUGE~\cite{rouge}, METEOR~\cite{meteor}: Text Similarity Evaluation Metrics}. 
After the introduction of VLMs/LLMs into the VAD task, many works have treated it as a VQA task or a video captioning task, providing corresponding textual annotations. As a result, the evaluation metrics from the NLP field have been introduced into the VAD domain.

In natural language processing (NLP) tasks, text similarity evaluation metrics are widely used to assess the quality of generated text compared to reference text: BLEU (Bilingual Evaluation Understudy) evaluates translation quality by calculating the precision of n-gram matches between the generated text and the reference text, focusing on surface-level similarity. ROUGE (Recall-Oriented Understudy for Gisting Evaluation) is a recall-oriented metric commonly used for text summarization, which measures the overlap of units (e.g., n-grams, word sequences) between the generated and reference summaries. METEOR (Metric for Evaluation of Translation with Explicit Ordering) combines precision, recall, and synonym matching to calculate a weighted harmonic mean score. It incorporates stemming and synonym libraries to enhance semantic understanding.
These metrics provide quantitative benchmarks for model optimization but come with their own limitations. For example, BLEU is sensitive to word order but neglects semantics, ROUGE prioritizes recall but may overlook conciseness, and METEOR relies on external resources and has computational complexity. In tasks like video understanding, where subjectivity and flexibility are significant factors, the performance of these metrics is often suboptimal.

\textbf{Evaluation Metrics Based on LLMs and MLLMs}.
In the VAD, particularly in the video anomaly understanding domain, evaluation metrics based on LLMs and MLLMs are gradually emerging as novel and effective assessment methods. LLM-based evaluation metrics draw inspiration from text similarity evaluation methods in natural language generation tasks (such as ROUGE and BLEU) and leverage the powerful semantic understanding and generation capabilities of LLMs to more comprehensively measure the similarity between generated descriptions and ground truth annotations~\cite{suvad,hawk_tang2024hawk}. LLMs can capture subtle semantic differences, contextual associations, and logical coherence in text:
\begin{equation}
	Metric_{\rm LLM} = {\rm LLM}(Result, \mathcal{Y}_{test})
\end{equation}
However, LLM-based metrics face issues such as instability, slow computation speeds, and a heavy reliance on the quality of prompt design.

In contrast, MLLM-based evaluation metrics combine video content and textual descriptions, utilizing a multi-modal fusion approach to enhance the ability to recognize complex scenarios and more accurately capture the semantic relationships and contextual consistency between video anomalies and textual descriptions~\cite{cuva_du2024uncovering,ecva}:
\begin{equation}
 	Metric_{\rm MLLM} = {\rm MLLM}(Result, \mathcal{X}_{test}, \mathcal{Y}_{test})
\end{equation}
These metrics guide the model to understand the task requirements through carefully designed prompts and integrate visual information with textual content for comprehensive judgment. However, their effectiveness also heavily depends on the quality of prompt design and the efficiency of video feature extraction. Additionally, they come with high computational costs, requiring continuous optimization and adjustment for practical applications.

\section{Overall of the Methods in Video Anomaly Detection}

\begin{figure*}[h]
	\centering
	\includegraphics[width=\linewidth]{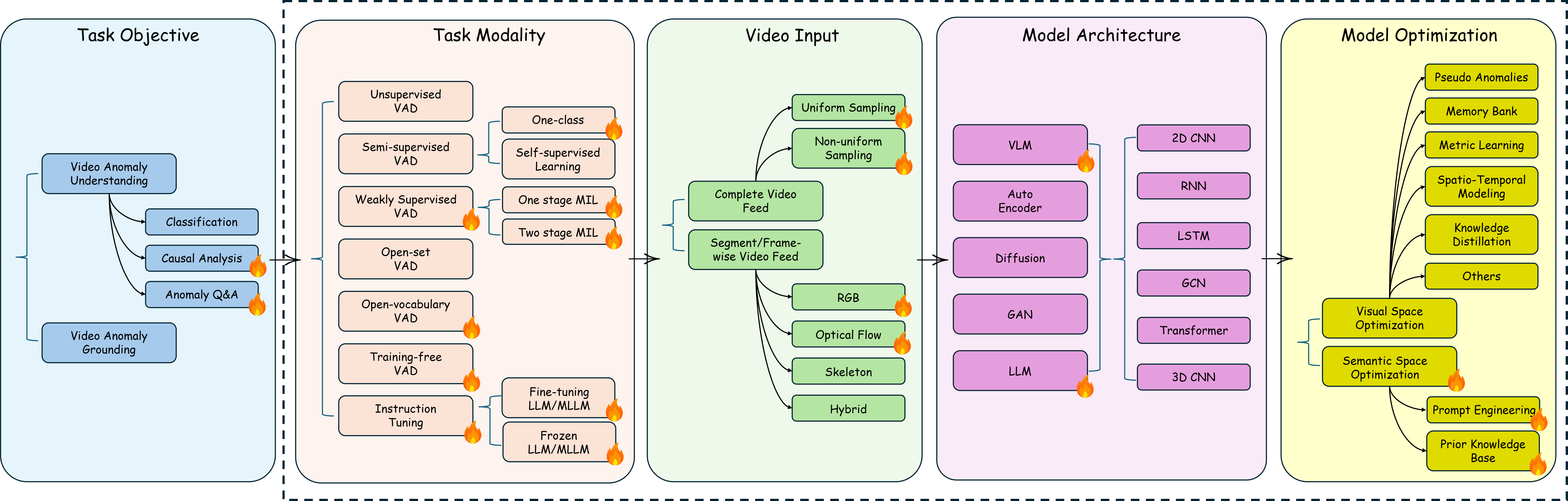}
	\caption{The framework tree we proposed that can be compatible with all existing types of VAD tasks. We break down VAD tasks into five components: Task Objective, Task Modality, Video Input, Model Architecture, and Model Optimization. In the figure, the flame icon indicates nodes where methods utilizing VLM/LLM have emerged. The Task Objective section has already been introduced earlier; below, we will elaborate on the branches rooted at Task Modality (the part within the dashed box).}
	\label{fig:main}
\end{figure*}

\subsection{overall}

Our in-depth investigation reveals that VAD methods employing VLMs/LLMs represent a fundamental paradigm shift compared to traditional methods based on DNNs. Directly classifying them together with traditional VAD methods clearly lacks theoretical justification. However, it is important to note that certain methods incorporating VLMs still retain the core characteristics of traditional paradigms at the task execution level. More significantly, some LLM/MLLM-based methods substantively inherit the key methodological principles of traditional research. For instance, HAWK~\cite{hawk_tang2024hawk} enhances motion feature extraction by introducing optical flow inputs, while VANE-Bench~\cite{vane-bench} adopts a pseudo-anomaly sample construction strategy to optimize the learning of classification boundaries.

Based on the above analysis, we argue that the introduction of VLMs/LLMs has fundamentally expanded and restructured the VAD domain. VAD methods based on large models exhibit a relative independence in research paradigm compared to traditional DNN-based methods, while also demonstrating significant methodological continuity and technological complementarity.

To systematically review the latest advancements in the field and construct a unified theoretical analysis framework, we proposes a novel classification framework with strong compatibility. The framework integrates five core dimensions: Task Objective, Task Modality, Video Input, Model Architecture and Model Optimization, establishing a unified analytical system for both traditional DNN-based methods and VLM/LLM-based methods (see Fig.~\ref{fig:main}). Notably, dimensions marked with a flame icon highlight the transformative impact of large-model technologies on the VAD domain.

In this section, we will first introduce the common task objectives of VAD. Then, we will provide a detailed overview of both classical and recent methods for different VAD tasks in separate subsections. Given that existing reviews have already provided comprehensive analyses of traditional DNN-based methods, we will focus on examining VAD methods based on VLMs/LLMs. Additionally, it will delve into the fundamental differences in technological pathways and theoretical frameworks between these methods and traditional methods.
\textbf{It is worth noting that we use ``$\star$'' to indicate sections related to VLMs/LLMs.}

\subsection{Task Objective}

\subsubsection{Video Anomaly Understanding}

As described in Section.~\ref{sec:analysis} and Section.\ref{sec:background}, video anomaly understanding primarily involves three core subtasks: Classification, Anomaly Q\&A, and Causal Analysis. The technical characteristics and research progress of these subtasks are summarized as follows:

\textbf{Classification}. 
This task is often jointly modeled with anomaly grounding, requiring models to identify anomalous video frames along the temporal dimension while accurately determining the category of the anomalous event. Since this task requires supervision with video-level classification labels, it is primarily applied within the framework weakly-supervised VAD. The predominant evaluation metric for this task is AP. On the XD-Violence benchmark dataset, early works~\cite{xd-violence,131-wu2022self,140-li2022self,160-zhang2023exploiting,162-wu2021learning,163-tian2021weakly,166-cho2023look,177-chen2023mgfn} achieved remarkable results by leveraging spatiotemporal feature extraction networks such as I3D~\cite{i3d} or VideoSwin~\cite{videoswin}. With the advancement of Vision-Language Models and Large Language Models, methods based on the CLIP architecture~\cite{161-yang2024text,175-lv2023unbiased,184-wu2024weakly,vadclip} have demonstrated superior generalization capabilities and performance in classification performance.

\textbf{Anomaly Question \& Answering}. 
This subtask represents a novel application paradigm of MLLMs in the domain of video anomaly detection and can be regarded as an extension of the Video Question Answering (Video QA) task in a specific domain. Current research primarily focuses on two paradigms: Training-free VAD and Instruction-tuned VAD based on pre-trained knowledge. Works such as LAVAD~\cite{lavad}, SUVAD~\cite{suvad}, VERA~\cite{vera_ye2025vera}, and AnomalyRuler~\cite{followtherules_yang2024follow} enable zero-shot anomaly reasoning by activating the inherent knowledge of the models. In contrast, the methods based on instruction tuning~\cite{184-wu2024weakly,vlavad,vad-llama,holmes-vau_zhang2025holmes} have significantly enhanced the fine-grained understanding capability of the question-answering task by constructing anomaly-related instruction datasets to fine-tune the models.

\textbf{Causal Analysis}. 
Although similar in form to anomaly question answering (both requiring answers to questions based on video content), this task imposes higher demands on the causal reasoning capabilities of models. It requires not only accurate descriptions of anomalous phenomena but also systematic analyses of the causes of anomalies and their potential impacts. Due to its emphasis on precise keyframe extraction and the construction of spatiotemporal causal chains, this task is significantly more challenging than conventional question-answering tasks. The CUVA~\cite{cuva_du2024uncovering} framework achieves keyframe localization through a learnable Multi-instance Spatiotemporal Attention Module (MIST~\cite{mist}) and establishes the first benchmark dataset for causal analysis. Subsequent research, such as ECVA~\cite{ecva}, further optimized spatiotemporal feature interaction mechanisms. At present, this field remains in its early stages of methodological exploration and evaluation system development.

\subsubsection{Video Anomaly Grounding}

Video Anomaly Grounding, as a mainstream and core task in the field of VAD, aims to precisely identify both the temporal intervals and the spatial distribution of anomalous events. 

As the predominant research direction within the task, the goal of temporal grounding is to distinguish anomalous segments from normal ones through frame-level anomaly score curves. The commonly used evaluation metric for this task is the AUC. Current methodologies encompass a variety of paradigms, including Unsupervised VAD, Semi-supervised VAD, Weakly-supervised VAD, Open-set VAD, Open-vocabulary VAD, Training-free VAD, and Instruction-tuned VAD. Classical methods and recent breakthroughs under each paradigm will be systematically elaborated in Section.~\ref{sec:sec:se} - Section.~\ref{sec:sec:opv}.

For spatial anomaly grounding, 
this task requires models to detect anomalies at the frame level while further grounding anomalous regions at the pixel level. It is typically evaluated using RBDC or TBDC. Early studies~\cite{tpami_6_ionescu2019object,tpami_25_ramachandra2020learning,futurepredict_liu2018future,street} conducted preliminary explorations based on handcrafted features or two-stage detection frameworks. Notably, Georgescu et al.~\cite{tpami_georgescu2021background} proposed a Background-Agnostic framework, which decouples scene semantics from anomalous motion patterns, achieving significant performance improvements in complex dynamic scenes. However, progress in further performance breakthroughs has been relatively limited due to the sparsity of anomalous data and the high cost of annotations.

\section{Semi-supervised VAD}
\label{sec:sec:se}

As previously discussed, SVAD only utilizes samples containing normal events during the training phase, making the traditional full supervised learning paradigm inapplicable. To address this issue, the most straightforward approach is to leverage the intrinsic information of the training samples to learn the patterns of normal events or, in other words, define the boundaries of normal events. Generally, the research paradigms for SVAD can be categorized into two main types: self-supervised learning and one-class classification.

\subsection{Paradigms}

\subsubsection{Self-Supervised Learning}
The core idea of self-supervised learning is to design auxiliary tasks to extract supervision signals from unlabeled data. In the SVAD task, where there is a severe imbalance between positive and negative samples and a lack of labeled anomalous samples, self-supervised learning becomes particularly crucial. It enables the learning of feature representations for normal samples, thereby establishing the patterns of normal behaviors. 

The self-supervised learning paradigm has consistently played a dominant role in SVAD tasks. The primary focus and challenge in this process lie in designing effective auxiliary tasks that allow the model to extract meaningful information from the limited normal samples. Currently, the commonly used forms of auxiliary tasks include the following.

\textbf{Reconstruction and Prediction}.
Reconstruction~\cite{convae_hasan2016learning,17_nguyen2019anomaly,22_wang2018generative,48_yu2023regularity,52_chang2020clustering,53_fang2020anomaly
,54_huang2021self,14_luo2017remembering} and prediction~\cite{futurepredict_liu2018future,26_liu2021hybrid,33_cai2021appearance,46_huang2022hierarchical,55_zhou2019attention,56_zhang2020normality,57_wang2021robust,58_yu2021abnormal,59_zhou2022object,60_cheng2023spatial,61_liu2023amp} are among the most common self-supervised tasks in the field of semi-supervised video anomaly detection (VAD). These approaches typically input a series of consecutive or evenly spaced video frames into the model and use the frames themselves as supervision signals. The distinction lies in their objectives: reconstruction aims to restore a specific frame within the input sequence, while prediction attempts to predict a frame following the input sequence.

During the reconstruction/prediction process, the model learns the features of video frames containing normal events, generating higher reconstruction/prediction errors for anomalous frames during inference. The larger the error, the higher the probability of the data being classified as anomalous. The optimization objective for this process can be formulated as:
\begin{equation}
	l_{recon} = \mathcal{P}(\Phi_{recon}(x_t,\theta),x_t),
\end{equation}
\begin{equation}
	l_{predict} = \mathcal{P}(\Phi_{predict}(x_{t-n}, ... , x_{t-2}, x_{t-1},\theta),x_t),
\end{equation}
where $n$ is the length of the input consecutive video frames. However, due to the strong generalization capability of deep neural networks, models may effectively fit unseen anomalous frames, even without prior exposure to them. Therefore, current research focuses on reducing the model's generalization ability to unseen anomalies through various optimization strategies.

\textbf{Video Frame Interpolation}.
Video frame interpolation~\cite{62_yu2020cloze,63_yang2023video,64_yu2023video} is inspired by the "fill-in-the-blank" training paradigm in natural language processing. In this auxiliary task, the model is provided with incomplete video sequences and is required to fill in the missing parts based on the context. Essentially, reconstruction/prediction can be regarded as a specific case of video frame interpolation. The optimization objective for this task can be expressed as:

\begin{equation}
	l_{inter} = \mathcal{P}(\Phi_{inter}(...,x_{t-1},x_{t+1},...,\theta),x_t).
\end{equation}

Because the missing parts in video sequences for interpolation tasks are not fixed, the model learns more robust spatiotemporal features during the completion process.

\textbf{Jigsaw Puzzles}.
Semi-supervised VAD methods using jigsaw puzzles as auxiliary tasks~\cite{65_wang2022video,66_shi2023video,67_barbalau2023ssmtl++} divide a series of video frames into image patches, shuffle their order, and require the model to reconstruct the original sequence. The optimization objective for this process is:

\begin{equation}
l_{jigsaw} = \sum_{t} \sum_{i} \mathcal{P}\big(\Phi_{jigsaw}(\text{shuffle}({p_{t,i}}), \theta), (t_{i})\big),
\end{equation}
where $p_{t,i}$ represents the $i$-th image patch in the $x_t$, and $t_{i}$ denotes the ground-truth spatiotemporal positional relationship of the image patches.

Compared to full-frame reconstruction/prediction tasks, jigsaw puzzle tasks explicitly learn both the temporal and spatial features of video segments. Additionally, because these methods do not require processing entire frames, they reduce the number of model parameters, improve computational efficiency, and capture richer spatiotemporal relationships.

\textbf{Contrastive Learning}.
In semi-supervised VAD, contrastive learning leverages the similarity and dissimilarity between samples to learn the features of normal samples. Specifically, if two samples are drawn from the same distribution, they are treated as positive pairs; otherwise, they are treated as negative pairs~\cite{68_huang2021abnormal}. For instance, Wang et al.~\cite{69_wang2020cluster} proposed a clustering attention contrastive framework for anomaly detection using contrastive learning. During inference, the highest similarity between a test sample and its different variants is used as the normality score. Lu et al.~\cite{70_lu2022learnable} further introduced a learnable locality-sensitive hashing (LSH) method that incorporates a contrastive learning strategy for anomaly detection.

\textbf{Denoising}.
Denoising tasks~\cite{49_flaborea2023multimodal,71_sun2020adversarial,72_chen2021nm} can be considered a variant of reconstruction tasks, with the key difference being the deliberate addition of noise to the input data. The optimization objective for this process can be formalized as:

\begin{equation}
	l_{denoise} = \mathcal{P}(\Phi_{denoise}(x_t + \eta, \theta), x_t),
\end{equation}
where $\eta$ is the noise added to the input.

During denoising training, the model learns more robust feature representations of normal samples, thereby enhancing its anomaly detection performance.

\textbf{Multi-Task Learning}.
Multi-task learning integrates multiple auxiliary tasks to learn more robust feature representations. Recent studies have explored jointly training VAD models on multiple pretext tasks. For example, several works have combined reconstruction and prediction~\cite{15_zhao2017spatio,27_bao2022hierarchical,39_morais2019learning,81_ye2019anopcn}, prediction and denoising~\cite{49_flaborea2023multimodal,82_liu2022appearance,83_liu2022learning}, prediction and jigsaw puzzles~\cite{84_huang2022self}, and prediction with contrastive learning~\cite{70_lu2022learnable}. Furthermore, other studies~\cite{66_shi2023video,67_barbalau2023ssmtl++,86_zhang2024multi} have focused on developing more sophisticated multi-task learning approaches.

Although multi-task learning may introduce higher computational overhead, the combination of complementary tasks can significantly improve the model's anomaly detection performance.

\subsubsection{One-Class Classification Learning}

Due to the lack of labeled anomalous samples, another approach for SVAD focuses on normal samples, assuming that they follow a specific distribution. By learning to fit this distribution, the model identifies samples that deviate significantly from it as anomalies. This type of method eliminates the need for complex auxiliary task design. It is worth noting that the design of self-supervised auxiliary tasks and one-class classification learning both aim to model the decision boundary of normal samples. As a result, their implementations may overlap in certain cases.

In traditional DNN-based methods, mainstream one-class classification approaches typically include Gaussian classifiers, adversarial classifiers based on GAN, and various one-class classifiers. With the development of VLMs or LLMs, which can map visual features to semantic features, more methods have begun to leverage pre-trained VLMs to extract semantic features of normal videos and construct classification boundaries in the semantic feature space. Compared to the visual feature space, the semantic space offers stronger generalization capabilities and explicitly incorporates semantic information. Consequently, VLM-based semi-supervised methods often achieve higher performance and demonstrate superior scene generalization.

\textbf{Gaussian Classifiers}.
Gaussian classifier-based methods~\cite{21_sabokrou2017deep,23_fan2020video,92_sabokrou2018deep} assume that normal samples follow a Gaussian distribution. During the training phase, the model learns the Gaussian distribution parameters (mean $\mu$ and variance $\sigma$) of normal samples. During the testing phase, samples that deviate significantly from the mean of this distribution are classified as anomalies.

\textbf{Adversarial Classifiers}.
Adversarial classifiers utilize GANs by leveraging the adversarial process between a generator $G$ and a discriminator $D$ to model the distribution of normal samples. Specifically, during training, the generator $G$ attempts to reconstruct normal samples with added Gaussian noise, while the discriminator $D$ determines whether the outputs of $G$ conform to the distribution of normal samples. In this framework, the anomaly score of a given test sample is derived from $D(G(x))$. To mitigate the instability of GAN training, Zaheer et al.~\cite{96_zaheer2020old,97_zaheer2022stabilizing} modified the task objective to differentiate between high-quality and low-quality reconstructions and constructed pseudo-anomalous samples to assist training.

\textbf{One-Class Classifiers}.
In the early stages of VAD research, one-class classifiers mainly included One-Class Support Vector Machine (OC-SVM)~\cite{87_scholkopf2001estimating} and Support Vector Data Description (SVDD)~\cite{88_tax2004support}. OC-SVM distinguishes normal and abnormal samples by learning a maximum-margin hyperplane. AMDN~\cite{18_xu2015learning} uses an autoencoder to extract features of normal samples and classifies all normal modalities. Deep SVDD~\cite{90_ruff2018deep,91_liznerski2020explainable} assumes that all normal samples are distributed within a bounded set and aims to find the smallest hypersphere that encloses all normal samples. DeepOC~\cite{20_xu2015learning} proposed an end-to-end deep one-class classifier for anomaly detection.

The above methods all learn the classification boundary of normal samples in the visual feature space. However, compared to self-supervised task designs, these methods are more cumbersome and have limited boundary fitting capabilities. This is mainly because relying solely on visual features makes it difficult to learn the true essence that distinguishes anomalies—namely, semantic-level features. In addition, differences in distribution between the training and test sets, the small proportion of abnormal content in video frames, and susceptibility to background interference all significantly affect model performance.

In recent years, VLM-based one-class classification methods have shown significant advantages.VLMs have already established connections between visual and semantic features during the pre-training stage. Compared to the visual feature space, the semantic feature space is less sensitive to scene and appearance changes. For example, both "riding a bike on the street in red clothes" and "riding a bike on a zebra crossing in a raincoat" can be mapped to the semantic feature of "riding a bike." Therefore, VAD methods based on VLMs are inherently more robust and generalizable. More importantly, under explicit semantic guidance, the classification boundary learned by the model is more precise.

For example, Gao et al.~\cite{accv} combines pre-trained action feature extraction models with clustering models with learnable prompts, which not only improves anomaly detection performance, but also allows for flexible adjustment of anomaly definitions through text guidance during testing. Other studies~\cite{doshi2023towards}, though not directly using VLMs, achieve interpretable anomaly detection supporting cross-domain adaptation by combining object detection networks with semantic embedding networks to compose video frame content in the semantic dimension.

\subsection{Video Input}

Due to the lack of abnormal sample annotations, existing SVAD methods find it difficult to directly process complete video sequences. A common practice is to divide videos into fixed-length segments~\textbf{(Segment/Frame-wise Video Feed)} and feed them into models in the form of RGB images, optical flow, skeletons, or hybrid inputs.

\subsubsection{RGB}
RGB images are the most commonly used input form in VAD tasks, with the advantage of requiring no additional preprocessing. According to different input methods, RGB images can be categorized into three types: frame-level RGB input, patch-level RGB input, and object-level RGB input. These images are usually stacked along the color channel or temporal dimension to capture both spatial and temporal information.

Frame-level input provides a global view of the entire scene, including both the background and the foreground where anomalies may occur. For example, methods such as ConvAE~\cite{convae_hasan2016learning}, ConvLSTM-AE~\cite{140-li2022self}and STAE~\cite{15_zhao2017spatio} use multi-frame RGB input, while AnomalyGAN~\cite{16_ravanbakhsh2017abnormal} and AMC~\cite{17_nguyen2019anomaly} focus on single-frame input.

Patch-level input focuses on local information by spatially or spatiotemporally slicing the frame-level RGB images, thereby reducing background interference. This approach helps to locate regions where anomalies are more likely to occur and improves detection accuracy. Representative methods include AMDN~\cite{18_xu2015learning,19_xu2017detecting}, DeepOC~\cite{20_xu2015learning}, and Deep-cascade~\cite{21_sabokrou2017deep} and GM-VAE~\cite{23_fan2020video}.

Object-level RGB input uses object detectors to extract foreground targets, almost completely removing background information, but may also lose the relationship between foreground and background. Hinami et al.~\cite{24_hinami2017joint} first proposed the object-level RGB-based VAD method FRCN, followed by methods such as HF2-VAD~\cite{26_liu2021hybrid}, HSNBM~\cite{27_bao2022hierarchical}, BDPN~\cite{28_chen2022comprehensive}, ER-VAD~\cite{29_sun2022evidential}, and HSC~\cite{30_sun2023hierarchical}. Object-level RGB input can significantly reduce background interference and better leverage prior knowledge from VLM pretraining, so most VLM-based semi-supervised VAD methods adopt this approach. For example, Gao et al.~\cite{accv} uses an object detector to determine the target position in the central frame and takes the corresponding regions from multiple preceding and following frames as input; Doshi et al.~\cite{doshi2023towards} feeds all targets within the same frame into the model together.

\subsubsection{Optical Flow}
Compared to static images, videos contain richer temporal contextual information, which helps to recognize events that cannot be detected from a single frame. Optical flow, as an important representation of motion information, is often used as an auxiliary input for VAD. Usually, optical flow is not used alone, but combined with RGB images to form a two-stream network. Optical flow input can also be divided into frame-level~\cite{futurepredict_liu2018future,31_luo2021future,32_wu2020fast}, patch-level~\cite{20_xu2015learning,36_li2020spatial,37_ramachandra2020learning}, and object-level~\cite{26_liu2021hybrid,35_huang2023video,38_reiss2022attribute} types.

\subsubsection{Skeleton}
Similar to object-level RGB and optical flow, skeleton information is also helpful for understanding behavioral dynamics in videos. Skeleton input focuses on the human body and is mainly used for detecting human-related abnormal events. Morais et al.~\cite{39_morais2019learning} first proposed learning normal human behavior patterns through dynamic skeletons. Subsequent methods such as GEPC, MTTP~\cite{41_rodrigues2020multi}, NormalGraph~\cite{42_luo2021normal}, HSTGCNN~\cite{43_zeng2021hierarchical}, TSIF~\cite{44_yang2022two}, STGCAELSTM~\cite{45_li2022human}, STGformer~\cite{46_huang2022hierarchical}, MoCoDAD~\cite{49_flaborea2023multimodal}, and TrajREC~\cite{50_stergiou2024holistic} all use skeletons as the foundation for human-centric VAD tasks.

\subsubsection{Hybrid}
Thanks to the complementarity of different modalities, hybrid inputs usually improve VAD performance more than single modalities. In deep learning-driven VAD methods, hybrid input has become a common practice. Typical hybrid forms include combining frame-level RGB with optical flow~\cite{futurepredict_liu2018future}, patch-level RGB with optical flow~\cite{20_xu2015learning}, and object-level RGB with optical flow~\cite{26_liu2021hybrid}. In addition, recent research has explored the use of multimodal inputs combining RGB and skeletons~\cite{51_pi2024eogt}.

\subsection{Model Architecture}

\subsubsection{Auto Encoder}
The auto encoder is one of the most commonly used network architectures in SVAD, consisting of two components: an encoder and a decoder. The encoder is responsible for compressing the input samples into latent feature representations, while the decoder reconstructs the original input from these latent representations. Auto-encoders are widely employed in self-supervised pre-training tasks based on image restoration, such as reconstruction~\cite{convae_hasan2016learning}, prediction~\cite{15_zhao2017spatio}, and inpainting~\cite{80_ristea2024self}. In addition, auto encoders are utilized in one-class learning-based methods~\cite{20_xu2015learning}, where the extracted features can further optimize subsequent one-class classifiers. The two main modules of the auto-encoder can be flexibly implemented based on various backbone network architectures, including 2D CNN~\cite{convae_hasan2016learning}, 3D CNN~\cite{15_zhao2017spatio}, RNN~\cite{76_luo2019video}, LSTM~\cite{14_luo2017remembering,101_medel2016anomaly}, GCN~\cite{42_luo2021normal,45_li2022human}, and Transformer architectures~\cite{63_yang2023video,80_ristea2024self}, among others.

\subsubsection{GAN}
As a powerful generative model, GAN has also been widely adopted in SVAD. Its core idea is to identify anomalous samples deviating from the normal mode through adversarial training between a generator $G$ and a discriminator $D$. GANs are frequently used in self-supervised learning paradigms for reconstruction or prediction tasks~\cite{futurepredict_liu2018future,102_ravanbakhsh2019training,105_feng2021convolutional}. Moreover, some one-class learning-based approaches~\cite{93_sabokrou2018adversarially,96_zaheer2020old} leverage the discriminator to estimate the 
probability that an input sample belongs to real data, with lower probabilities indicating a higher likelihood of anomaly.

\subsubsection{Diffusion Model}
As an emerging and powerful class of generative models, diffusion models have also garnered increasing attention in the VAD research community. Diffusion models learn the distribution of normal samples by progressively “denoising” and reconstructing samples. Similar to GANs, diffusion models are applied to self-supervised learning paradigms for reconstruction and prediction tasks. Yan et al.~\cite{7_yan2023feature} and Flaborea et al.~\cite{49_flaborea2023multimodal} proposed novel diffusion model-based methods, utilizing RGB frames and skeletal features as inputs, respectively, for the detection of anomalous events in videos.

\subsubsection{VLM}
As a type of multimodal pre-trained model, VLMs are capable of capturing high-level semantic information in videos by pre-learning the associations between visual and linguistic features. Compared to traditional approaches that rely solely on visual features, VLMs can establish more robust classification boundaries in the semantic space, thereby better adapting to scene changes and enhancing model generalization. For example, Gao et al.~\cite{accv} utilizes CLIP~\cite{clip} to map a series of object RGB features into the semantic space and directly identifies anomalous samples via clustering models.

\subsection{Model Optimization}

Due to the lack of abnormal annotations in SVAD, it is challenging for researchers to optimize models in the semantic space. As for optimization strategies within the visual space, the current approaches primarily include pseudo anomalies, memory banks, and knowledge distillation.

\subsubsection{Pseudo Anomalies}
In SVAD, the task is defined such that no anomaly-related annotations are used during training. Therefore, constructing pseudo-anomalies to balance positive and negative samples is a natural and effective strategy. Currently, methods for generating pseudo-anomalies can be categorized into three main types: (1) Perturbing normal samples, which involves introducing random perturbations to normal video samples, such as adding noise, shuffling frame sequences, or inserting additional image patches~\cite{astrid2021learning,106_wu2023dss,107_astrid2022limiting,109_astrid2023pseudobound}; (2) Utilizing generative models, where samples that resemble normal data but possess anomalous characteristics are generated using GAN or diffusion models~\cite{97_zaheer2022stabilizing,110_pourreza2021g2d}; (3) Simulating specific anomalous behaviors, i.e., manually introducing anomalous samples at the image or feature level~\cite{tpami_georgescu2021background,112_liu2023generating}. It is important to note that the constructed pseudo-anomalies should generally resemble the types of anomalies present in the test set. Although this approach may raise concerns regarding “data leakage,” the introduction of pseudo-anomalies does help the model learn a broader range of anomalous features, thereby improving detection performance.

\subsubsection{Memory Banks}
Memory banks~\cite{113_leng2022anomaly,114_yu2022effective,115_liu2023diversity,116_liu2023stochastic,118_wang2023memory} are used to store feature representations of normal video samples. These features serve as abstract representations of normal samples and can be dynamically updated to accommodate new normal patterns. This helps the model learn normal modalities more effectively while reducing confusion caused by overfitting. Memory banks have been applied in various paradigms, such as reconstruction (or prediction)~\cite{park2020learning,121_lv2021learning,122_yang2022dynamic} and contrastive learning~\cite{123_cao2024context}.

\subsubsection{Knowledge Distillation}
As a model compression technique, knowledge distillation can significantly improve inference speed. Ristea et al.~\cite{80_ristea2024self} methods employ a teacher-student network architecture, where the student network learns feature representations of normal videos. This not only enables extremely high inference speeds, but the discrepancies between the outputs of the teacher and student networks can also be mutually informative, thereby enhancing detection performance.

\subsection{Performance Comparison and Paradigm Example}

\begin{figure}[h]
	\centering
	\includegraphics[width=\linewidth]{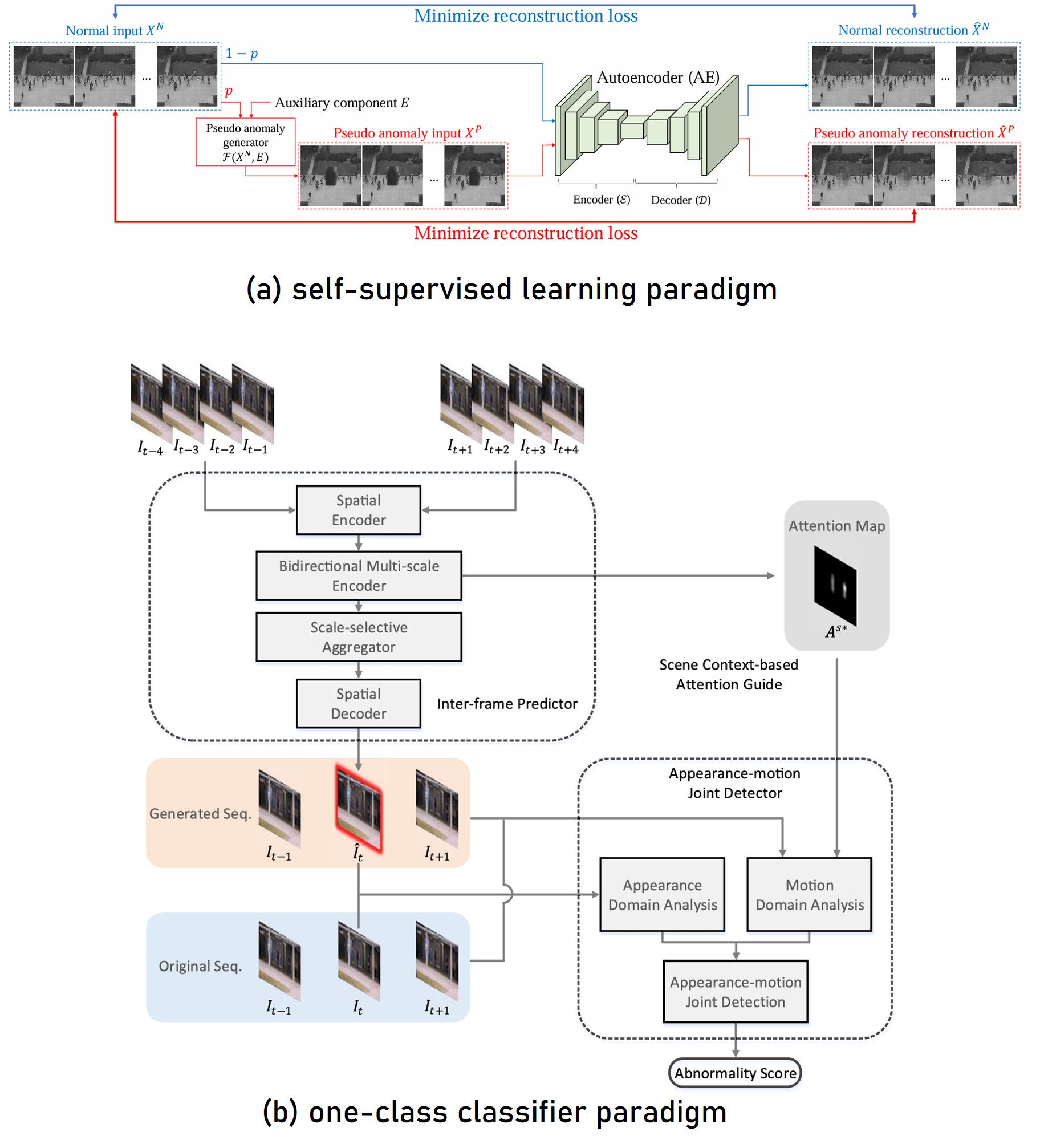}
	\caption{Fig (a) illustrates the reconstruction paradigm with synthetic pseudo-anomalies (LNRA~\cite{astrid2021learning}) in the self-supervised learning framework, while Fig (b) shows the one-class classifier paradigm (BMAN~\cite{124_lee2019bman}) in one-class classification learning.}
	\label{fig:svad}
\end{figure}

We provide a performance comparison of existing methods in Table~\ref{table:svad}, and present the classic paradigm of SVAD in Fig.~\ref{fig:svad}. Due to space limitations, we did not include the performance comparison of all methods in the table. Instead, we selectively present the most advanced methods and highly cited classic methods.

\begin{table}
	\caption{Comparison of the performance of existing SVAD methods}
	\label{table:svad}
	
	\resizebox{\linewidth}{!}{
	\begin{tabular}{cc|cccc}
		\toprule
		\multirow{2}{*}{Method} & \multirow{2}{*}{Year} & \multicolumn{4}{c}{Dataset} \\
		\cmidrule(r){3-6}
		& & Ped2(AUC) & Avenue(AUC) & SHTech(AUC) & UBnormal(AUC) \\
		\midrule
		ConvAE~\cite{convae_hasan2016learning} & 2016 & 90.0\% & 72.0\% & / & / \\
		STAE~\cite{15_zhao2017spatio}& 2017 & 91.2\% & 80.9\% & / & / \\
		FutureFrame~\cite{futurepredict_liu2018future}& 2018 & 95.4\% & 85.1\% & 72.8\% & / \\
		MemAE~\cite{119_gong2019memorizing} & 2019 & 94.1\% & 83.3\% & 71.2\% & / \\
		BMAN~\cite{124_lee2019bman} & 2019 & 96.6\% & 90.0\% & 76.2\% & / \\
		MNaD~\cite{120_park2020learning} & 2020 & 97.0\% & 88.5\% & 70.5\% & / \\
		AmmcNet~\cite{33_cai2021appearance} & 2021 & 96.6\% & 86.6\% & 73.7\% & / \\
		MultiTVAD~\cite{85_georgescu2021anomaly} & 2021 & 99.8\% & 92.8\% & 90.2\% & / \\
		JigsawPuzzle~\cite{65_wang2022video} & 2022 & 99.0\% & 92.2\% & 84.3\% & 56.4\% \\
		BDPN~\cite{28_chen2022comprehensive} & 2022 & 98.3\% & 90.3\% & 78.1\% & / \\
		SenceAware~\cite{30_sun2023hierarchical} & 2023 & 98.1\% & 93.7\% & 83.4\% & / \\
		LERF~\cite{117_sun2023learning} & 2023 & 99.4\% & 91.5\% & 78.6\% & / \\
		Nomral-Pose VAD~\cite{47_hirschorn2023normalizing} & 2023 & / & / & 85.9\% & 71.8\% \\
		SDMAE~\cite{80_ristea2024self} & 2024 & 95.4\% & 91.3\% & 79.1\% & 58.5\% \\
		FGDVAD~\cite{tan2024frequency} & 2024 & / & 88.0\% & 78.6\% & 68.9\% \\
		VADMamba~\cite{lyu2025vadmamba} & 2025 & 98.5\% & 91.5\% & 77.0\% & / \\
		MA-PDM~\cite{zhou2025video} & 2025 & 98.6\% & 91.3\% & 79.2\% & 63.4\% \\	
		\bottomrule
	\end{tabular}
}
\end{table}

\section{Weakly Supervised VAD}
\label{sec:sec:we}
Among all conventional DNN-based VAD methods, weakly supervised VAD (WVAD) has consistently been a focal point of research within the academic community. Although research on WVAD began somewhat later than that on SVAD, its use of video-level anomaly annotations aligns more closely with real-world application scenarios. Without the need for extensive manual labeling efforts, WVAD can effectively achieve anomaly detection in complex environments and attain relatively satisfactory detection performance.

\subsection{Paradigm}

\subsubsection{One Stage MIL}

One Stage MIL is the most fundamental and widely used paradigm in WVAD. Its core idea is to identify the most likely anomalous segments in both normal and abnormal videos and require the model to distinguish between them. Specifically, the typical procedure involves dividing long videos into multiple segments, employing the MIL mechanism to select the most representative segment from each video, and maximizing the score difference between the ``most anomalous'' segment in normal and abnormal videos. In this way, the model’s predicted anomaly probability for normal segments gradually decreases, while the predicted anomaly probability for abnormal samples gradually increases, thereby achieving anomaly detection. 
The optimization objective for this process is:
\begin{equation}
	l_{osm} = \sum_{t} \mathcal{P}(1, 1 - \max \Phi_{osm}(x_t^{a},\theta) + \max \Phi_{osm}(x_t^{n},\theta)),
\end{equation}
where $x_t^{a}$ denotes video frames from abnormal videos, and $x_t^{n}$ denotes video frames from normal videos. Many studies~\cite{deepmil_sultani2018real,157_zhang2019temporal,158_liu2022collaborative} have adopted this paradigm.

Furthermore, the TopK~\cite{xd-violence} further refines the one-stage MIL paradigm by considering the temporal continuity of events. Instead of relying solely on the most anomalous segment, it computes the average anomaly score of the top-K video segments, thereby enhancing the model’s detection performance.

Recently, an increasing number of studies have sought to incorporate the pretrained knowledge of VLMs to further optimize the model’s discriminative space through semantic guidance. For example, VadCLIP~\cite{vadclip} designs a fine-grained dual-branch structure for language and vision based on CLIP, transferring CLIP’s pretrained knowledge to the task of WVAD. Other works, such as LPE~\cite{LPE_pu2024learning}, AnomalyCLIP~\cite{anomalyclip_zanella2024delving}, and TPWNG~\cite{161-yang2024text} introduce learnable prompts and CLIP models to enhance contextual modeling capabilities and semantic discriminability. LAP~\cite{lap_tao2024learn} generates pseudo-anomalous labels by analyzing the anomalous similarity between prompts and video subtitles, thereby guiding the model to identify potential anomalous events.

The main advantage of the one stage MIL lies in its simplicity and efficiency, which has led to its widespread adoption. However, because it relies solely on video-level labels, it often performs well in detecting obvious anomalies but tends to overlook subtle or less perceptible anomalous events.

\subsubsection{Two Stage MIL}

In contrast, two stage MIL extends the one stage MIL paradigm. Its core idea is to leverage the predictions of the base model trained in the one stage to automatically select high-confidence anomalous regions as pseudo-labels for model retraining, thereby achieving adaptive enhancement. This two stage training strategy effectively boosts the performance of models in WVAD, further improving their generalization ability, especially in detecting subtle anomalies. Methods such as NoiseCleaner~\cite{137_zhong2019graph}, MIST~\cite{159_feng2021mist}, MSL~\cite{140-li2022self} and CUPL~\cite{160-zhang2023exploiting} all adopt the typical two-stage MIL approach and have achieved promising results.

Although from a conceptual perspective, there seems to be little fundamental difference between two stage MIL and one stage MIL, the two paradigms have been recognized and distinguished by scholars in the field of VAD. Therefore, we list it here as a separate paradigm. Two stage MIL methods perform well in WVAD, but they also have disadvantages such as high computational complexity and label noise confusion. On one hand, both the pre-training and self-training stages involve multiple rounds of iterative training, resulting in high computational costs. On the other hand, the second stage relies heavily on the initial model generated during pre-training. If the quality of the initial model is poor, incorrect predictions may be taken as pseudo-labels, affecting the subsequent training performance.

\subsection{Video Input}

Unlike SVAD, WVAD does not require the design of self-supervised tasks such as reconstruction or prediction. As a result, most weakly supervised VAD methods first use pretrained models to extract visual features, which are then fed into the model. Using pretrained features as input can effectively leverage the appearance and motion knowledge learned by the pretrained models, significantly reducing the complexity of the detection model and enabling efficient training. However, similar to SVAD, weak supervision generally uses segmented video clips as input~\textbf{(Segment/Frame-wise Video Feed)}, since the provided annotations are also incomplete.

\subsubsection{RGB}

Similar to SVAD, RGB video segments represent the most common input modality for WVAD. A typical approach involves dividing a long video into multiple segments and extracting global features from each segment using pretrained visual models. For the selection of pretrained visual models, C3D has been employed in several studies~\cite{127_tran2015learning,128_zaheer2020claws,deepmil_sultani2018real}, while I3D has been widely adopted in other works~\cite{i3d,131-wu2022self,132_zhou2024batchnorm,133_almarri2024multi}.
Some studies have utilized 3DResNet~\cite{134_hara2018can,135_sun2023long}, whereas TSN has also been explored~\cite{138_li2022weakly}.

\subsubsection{Optical Flow}

Similar to semi-supervised VAD, motion-focused inputs such as optical flow have also attracted attention from researchers in weakly supervised VAD. However, due to the time-consuming nature of optical flow extraction, this modality is less frequently employed in existing approaches. Some studies have explored the use of optical flow information as input to achieve improved performance~\cite{137_zhong2019graph}.

\subsubsection{Audio}

Although not as commonly utilized as visual signals, audio, as a one-dimensional signal, often contains important perceptual information. Certain datasets, such as XD-Violence~\cite{xd-violence}, provide crucial audio signals. In several studies, including those employing VGGish~\cite{145_pang2021violence,146_peng2023learning}, audio signals are first resampled, followed by the computation of spectrograms and the creation of log-mel spectrograms. These features are then segmented into non-overlapping samples, which are subsequently fed into pretrained audio models to assist in anomaly detection.

\subsubsection{Text}

Since VLMs align semantic and visual information during pretraining, using text as input to guide model training is a natural approach. Recently, several works~\cite{LPE_pu2024learning,anomalyclip_zanella2024delving,161-yang2024text,vadclip,lap_tao2024learn} have explored the use of learnable text prompts to guide the reconstruction of decision boundaries in the visual space. This strategy has significantly improved the generalization ability of the models.

\subsubsection{Hybrid}

The combination of multiple input modalities can effectively compensate for the limitations of individual signals and thereby enhance model performance. Common multimodal inputs include RGB combined with optical flow~\cite{143_wan2020weakly}, RGB combined with audio~\cite{151_wei2022look,153_yu2022modality}, RGB combined with both optical flow and audio~\cite{154_wu2022weakly}, as well as the recent integration of RGB with text~\cite{vadclip,LPE_pu2024learning,lap_tao2024learn,uca}.

\subsection{Model Architecture}

Unlike SVAD methods, which typically feature clearly defined architectures such as auto-encoders, GANs, or diffusion models, WVAD models are generally designed for the extraction of video features.

The model architectures in WVAD have evolved from early 3D CNNs (such as C3D and I3D)~\cite{deepmil_sultani2018real} to temporal modeling approaches (including LSTM and TCN)~\cite{127_tran2015learning,128_zaheer2020claws,131-wu2022self,132_zhou2024batchnorm,133_almarri2024multi,135_sun2023long}, graph-based structures (such as GCN)~\cite{140-li2022self}, and more recently to self-attention mechanisms (Transformers)~\cite{168_huang2022weakly,169_zhang2022weakly,170_zhou2023dual}.
Compared to other architectures, 3D CNNs are effective for capturing local spatio-temporal features, LSTM and TCN excel at modeling sequential dependencies, and GCNs are well-suited for representing relationships between entities (e.g., skeletons or dynamic graphs). Transformers, with their ability to model global dependencies, have emerged as one of the most promising architectures in this field.

With the introduction of pretrained vision-language models such as CLIP~\cite{clip}, the field of VAD is rapidly advancing toward a new paradigm characterized by unified representations, multi-task learning, and semantic guidance~\cite{LPE_pu2024learning,vadclip,lap_tao2024learn,tsa_clip,anomalyclip_zanella2024delving,161-yang2024text}. These models enable the integration of rich semantic information from both visual and textual modalities, facilitating the development of frameworks that can jointly address multiple tasks while leveraging shared representations. As a result, VAD research is increasingly focused on designing systems that not only detect anomalies but also benefit from semantic alignment and cross-modal understanding.

\subsection{Model Optimization}

\subsubsection{Visual Space Optimization}

\textbf{Spatio-Temporal Modeling}.
Anomalous events often manifest as localized spatiotemporal disruptions; therefore, mainstream approaches emphasize the joint modeling of spatial and temporal features. On the one hand, temporal modeling is employed to capture dynamic changes between segments; on the other hand, spatial modeling aims to pinpoint the precise regions where anomalies occur. Relevant techniques include: Temporal Convolutional Networks (TCN), which utilize one-dimensional or multi-dimensional convolutions to extract long-range temporal dependencies~\cite{130_wu2020not,164_liu2022decouple}; dilated convolutions, which expand the receptive field and enhance the perception of spatial context for anomalies~\cite{163-tian2021weakly}; Graph Convolutional Networks (GCN), which model spatial neighborhood relationships and are well-suited for capturing spatial dependencies in complex scenarios~\cite{166-cho2023look,165_chang2021contrastive}; Conditional Random Fields (CRF), which optimize spatial label consistency and improve the accuracy of spatial segmentation~\cite{167_purwanto2021dance}; and Transformers, whose self-attention mechanisms facilitate the modeling of long-range temporal and spatial dependencies, thereby significantly enhancing the detection of complex anomalies~\cite{140-li2022self,168_huang2022weakly,169_zhang2022weakly,170_zhou2023dual}.

\textbf{Knowledge Distillation}.
Knowledge distillation is a model compression and transfer learning technique that is frequently employed in anomaly detection, particularly in multimodal scenarios. Through distillation, the knowledge embedded in an information-rich teacher model (e.g., multimodal branches) is transferred to a student model (e.g., unimodal branches), thereby enhancing the latter’s detection performance under conditions of scarce or missing modality information~\cite{153_yu2022modality,182_liu2023distilling}. In addition, knowledge distillation can facilitate model lightweighting, making it suitable for deployment on edge devices.

\textbf{Metric Learning}
Although MIL-based classification ensures inter-class separability of features, such separability at the video level alone is insufficient for accurate anomaly detection. In contrast, enhancing feature discriminability by clustering similar features and isolating dissimilar ones should supplement or even strengthen the separability achieved by MIL-based classification. Specifically, the fundamental principle of feature metric learning is to make similar features compact in the feature space while pushing dissimilar features far apart, thereby improving discriminative power. At present, numerous studies~\cite{132_zhou2024batchnorm,162-wu2021learning,168_huang2022weakly,180_fioresi2023ted,181_zaheer2023clustering} have leveraged feature metric learning to enhance feature discriminability.

\subsubsection{Semantic Space Optimization}

\textbf{Prompt Engineering}
With the widespread adoption of vision-language models such as CLIP, prompt engineering has emerged as an innovative approach for optimizing the semantic space. In the field of WVAD, several methods utilizing pre-trained vision-language models have explored the use of learnable prompts to enable better identification of anomalous samples within the vision-language aligned space. Approaches such as CLIP-TSA~\cite{tsa_clip}, VadCLIP~\cite{vadclip}, LPE~\cite{LPE_pu2024learning}, ~\cite{anomalyclip_zanella2024delving} and Yang et al.~\cite{161-yang2024text} have designed specific anomaly description templates to guide the model’s attention toward anomalous event features. This strategy leverages the knowledge base of large models to improve discriminative power and offers enhanced generalization capability.

\subsection{Performance Comparison and Paradigm Example}

\begin{figure}[h]
	\centering
	\includegraphics[width=\linewidth]{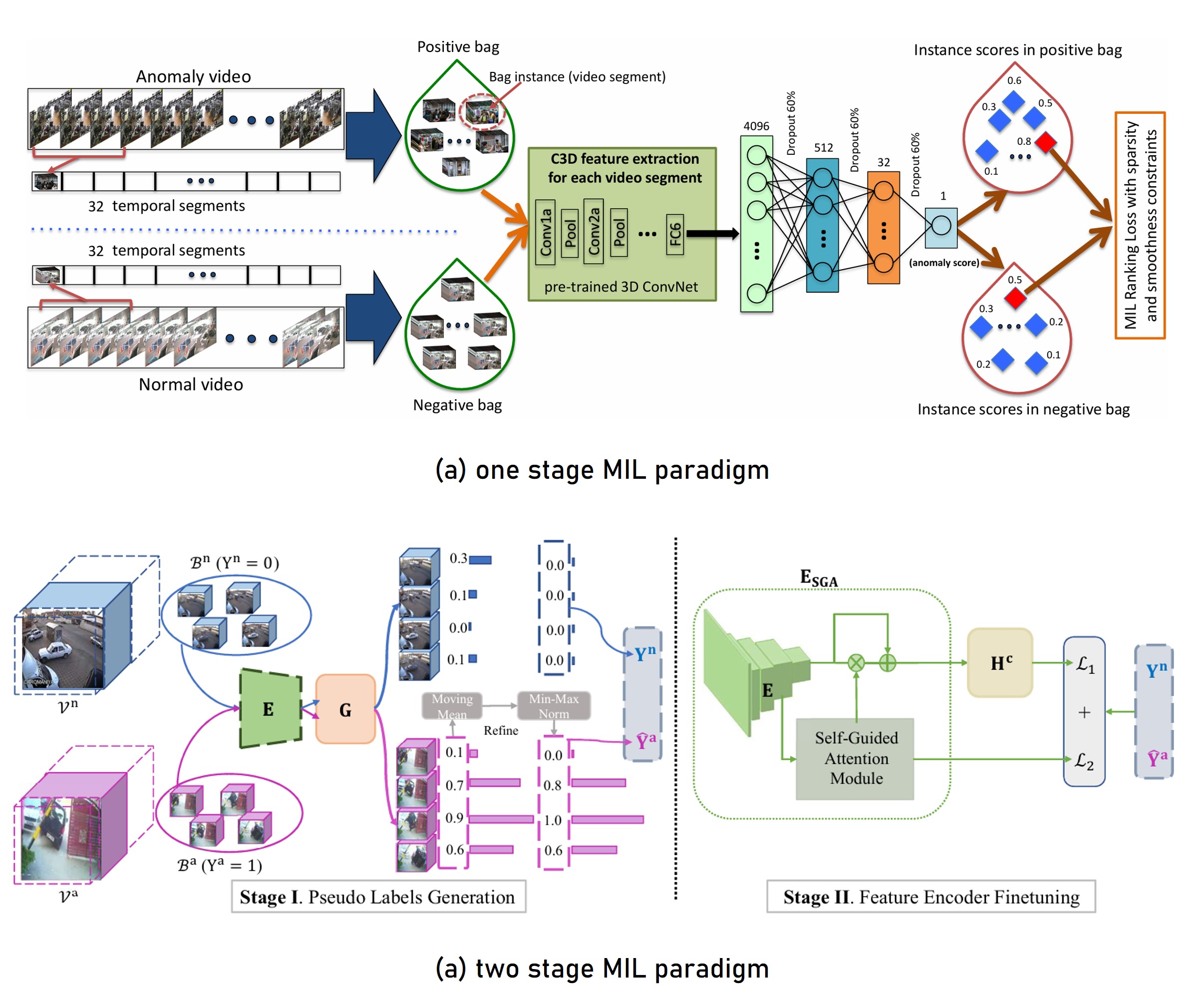}
	\caption{Fig (a) illustrates the one stage MIL paradigm (DeepMIL~\cite{deepmil_sultani2018real}), while Fig (b) shows the two stage MIL paradigm (MIST~\cite{159_feng2021mist}).}
	\label{fig:wvad}
\end{figure}

We provide a performance comparison of existing methods in Table~\ref{table:wvad}, and present the classic paradigm of WVAD in Fig.~\ref{fig:wvad}. Due to space limitations, we did not include the performance comparison of all methods in the table. Instead, we selectively present the most advanced methods and highly cited classic methods.

\begin{table}
	\caption{Comparison of the performance of existing WVAD methods}
	\label{table:wvad}
	
	\resizebox{\linewidth}{!}{
		\begin{tabular}{cc|ccc}
			\toprule
			\multirow{2}{*}{Method} & \multirow{2}{*}{Year} & \multicolumn{3}{c}{Dataset} \\
			\cmidrule(r){3-5}
			&  & UCF-Crime(AUC) & XD-Violence(AUC) & SHTech(AUC) \\
			\midrule
			
			DeepMIL~\cite{deepmil_sultani2018real} & 2018 & 75.40\% & / & / \\
			GCN~\cite{137_zhong2019graph} & 2019 & 82.12\% & / & 84.44\% \\
			CLAWS~\cite{128_zaheer2020claws} & 2020 & 83.03\% & / & 89.67\% \\
			MIST~\cite{159_feng2021mist} & 2021 & 82.30\% & / & 94.83\% \\
			MSL~\cite{140-li2022self} & 2022 & 85.62\% & 78.59\% & 97.32\% \\
			S3R~\cite{131-wu2022self} & 2022 & 85.99\% & 80.26\% & 97.48\% \\
			UMIL~\cite{175-lv2023unbiased} & 2023 & 86.75\% & / & / \\
			DMU~\cite{170_zhou2023dual} & 2023 & 86.97\% & 81.66\% & / \\
			VadCLIP~\cite{vadclip} & 2024 & 88.02\% & 84.51\% & / \\
			CLIP-TSA~\cite{tsa_clip} & 2024 & 87.58\% & 82.19\% & 98.32\% \\
			CMFVAD~\cite{cmfvad} & 2024 & / & 86.34\% & / \\

			\bottomrule
		\end{tabular}
}
\end{table}

\section{Unsupervised VAD}
\label{sec:sec:un} 

Compared to other supervised VAD methods, unsupervised VAD~(UVAD) offers the advantages of eliminating the need for labor-intensive manual annotations and enabling the model to autonomously determine anomaly boundaries. On the one hand, it is difficult to clearly define what constitutes normal human behavior in real-world scenarios; for instance, riding a bicycle on a playground may be considered acceptable, whereas riding in a corridor is often deemed anomalous. On the other hand, it is impractical to anticipate all possible normal events in advance, especially in real-world applications. Therefore, early research into UVAD held significant academic value. However, with the advent of vision-language models and large language models that provide rich pre-trained semantic information, the advantages of UVAD have diminished. Currently, the progress of research in UVAD has been relatively slow.

\subsection{Paradigms}
The complete absence of annotation information endows UVAD with substantial flexibility. Moreover, due to the limited research in this area, a standardized paradigm has yet to be established. The core principle of UVAD is to exploit the assumption that anomalous events occur far less frequently than normal events, drawing on or integrating paradigms from SVAD and WVAD to mine rare anomalous samples from large volumes of data.

\subsection{Video Input}

The way of using complete video input is rarely discussed in current research on UVAD. In contrast, segment-based or frame-by-frame input methods are more commonly adopted. Given the higher task complexity and limited research in UVAD compared to other VAD settings, most existingUVAD methods primarily utilize RGB frames from videos as model inputs. The basic strategy adopted by Ionescu et al.~\cite{208_tudor2017unmasking}, Lin et al.~\cite{209_liu2018classifier}, Lin et al.~\cite{213_lin2022causal}, and Wang et al.~\cite{210_wang2018detecting} involves segmenting videos and using these segments as model inputs. Modalities such as optical flow, skeleton data, and hybrid input forms have not yet been fully explored in current studies, which may represent promising directions for future research.

\subsection{Model Architecture}

\subsubsection{Auto-Encoder}
The auto-encoder is one of the most commonly used network architectures in UVAD, aiming to emulate the prediction/reconstruction paradigm of SVAD. Wang et al. [210] proposed a two-stage training mechanism, in which an auto-encoder is first trained with an adaptive reconstruction loss threshold to estimate normal events, followed by training an OC-SVM using pseudo-labels to further refine the normality model. Hu et al.~\cite{212_hu2022detecting} utilized masked auto-encoders, leveraging the rarity of anomalous events and the resulting prediction errors to achieve anomaly detection and scoring. Building on this, Yu et al.~\cite{214_yu2022deep} introduced an adaptive stepwise optimization strategy that combines deep reconstruction with localization-based reconstruction, significantly improving detection performance. In addition, Li et al.~\cite{218_li2021deep} combined clustering techniques with auto-encoders, iteratively filtering normal candidate samples based on reconstruction error, providing a new perspective for UVAD.

\subsubsection{GAN}
In addition to auto-encoders, GAN have also been applied in UVAD. Zaheer et al.~\cite{215_zaheer2022generative} proposed an unsupervised generative cooperative learning approach, which significantly improves anomaly detection performance by leveraging cross-supervision between the generator and discriminator, as well as the low-frequency nature of anomalies.

\subsubsection{Others}
In addition to the aforementioned architectures, Lin et al.~\cite{213_lin2022causal} proposed a causal inference framework characterized by a unique reasoning mechanism distinct from traditional model architectures. This approach integrates long-term temporal context with local image context to mitigate the impact of noisy pseudo-labels on anomaly detection.

\subsection{Model Optimization}

Currently, there has been no attempt to introduce semantic information in the field of UVAD. As a result, existing methods focus on optimizing models within the visual feature space.

\subsubsection{Pseudo Anomalies}
In terms of pseudo-label generation and optimization, Wang et al.~\cite{210_wang2018detecting} adopted a two-stage approach in which pseudo-labels are generated by marking normal events based on the reconstruction performance of an auto-encoder, and these pseudo-labels are then used to refine the normality model. Pang et al.~\cite{211_pang2020self} proposed a self-training deep ordinal regression method, utilizing classical one-class algorithms to generate initial pseudo-labels and subsequently iteratively optimizing the anomaly detector. Al-lahham et al.~\cite{216_al2024coarse} designed a coarse-to-fine pseudo-label generation framework that combines hierarchical clustering and statistical hypothesis testing to generate pseudo-labels, achieving significant results at both the video and segment levels.

\subsubsection{Spatio-Temporal Modeling}
In terms of spatio-temporal modeling, Ionescu et al.~\cite{208_tudor2017unmasking} introduced the "unmasking" technique, capturing anomalies by analyzing differences in classifier performance between consecutive events. Subsequently, Liu et al.~\cite{209_liu2018classifier} further improved this approach by integrating it with multi-classifier two-sample tests from statistical machine learning, significantly enhancing detection performance.

\subsubsection{Others}
In addition to the aforementioned optimization methods, Yu et al.~\cite{214_yu2022deep} proposed an adaptive stepwise optimization strategy that significantly improves detection performance by progressively refining the reconstruction process. This strategy integrates the characteristics of localization-based reconstruction and hybrid optimization, offering a new perspective for visual space optimization. Furthermore, Lin et al.~\cite{213_lin2022causal} reduced the impact of noisy pseudo-labels on the optimization process through a causal inference framework.

\subsection{Performance Comparison and Paradigm Example}

\begin{figure}[h]
	\centering
	\includegraphics[width=\linewidth]{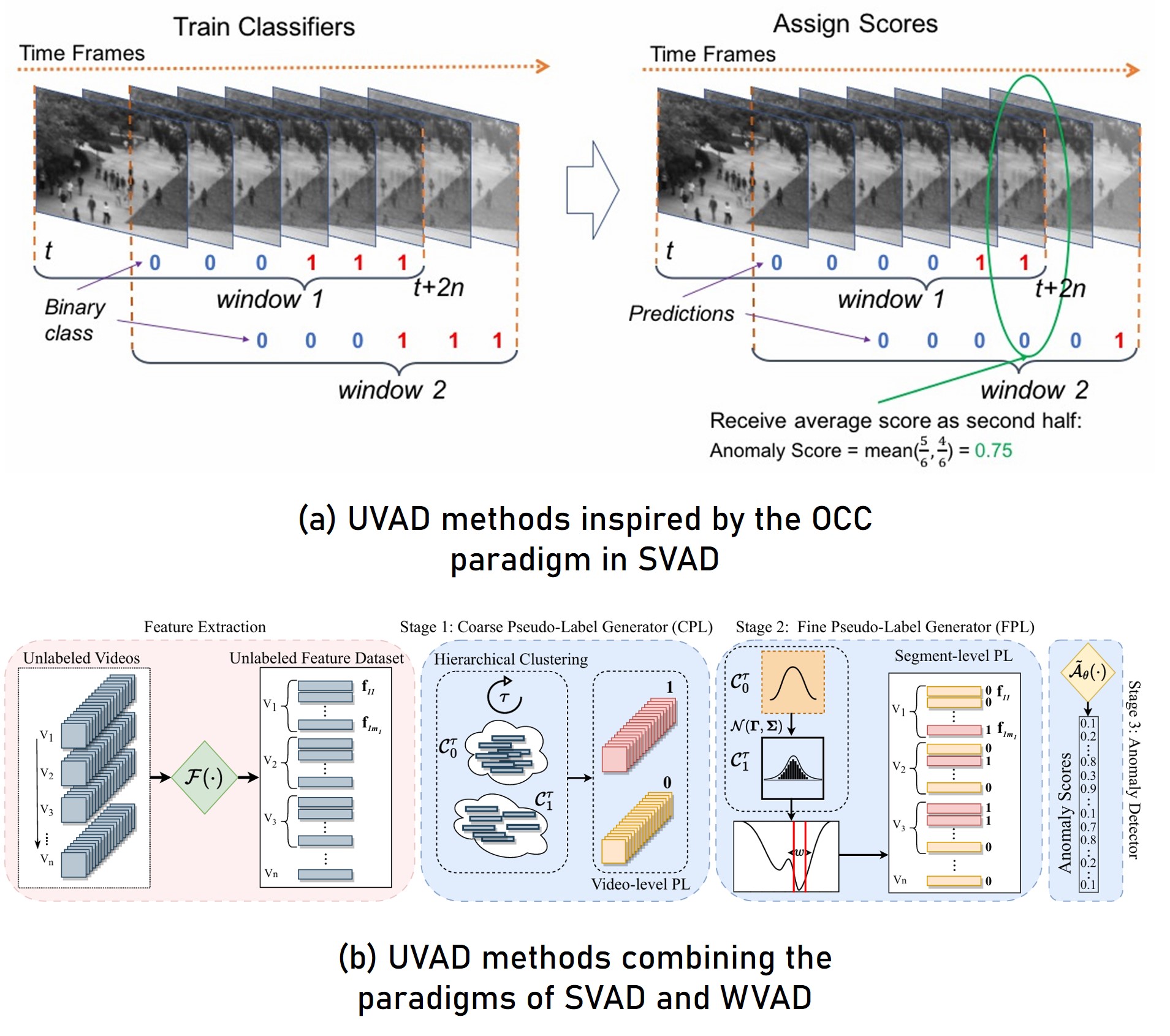}
	\caption{Fig (a) illustrates the method inspired by the OCC paradigm (MC2ST~\cite{209_liu2018classifier}), while Fig (b) shows the method combining the paradigms of SVAD and WVAD (C2FPL~\cite{216_al2024coarse}).}
	\label{fig:uvad}
\end{figure}

We provide a performance comparison of existing methods in Table~\ref{table:uvad}, and present the classic paradigm of WVAD in Fig.~\ref{fig:uvad}. Due to space limitations, we did not include the performance comparison of all methods in the table. Instead, we selectively present the most advanced methods and highly cited classic methods.

\begin{table}
	\caption{Comparison of the performance of existing UVAD methods}
	\label{table:uvad}
	
	\resizebox{\linewidth}{!}{
		\begin{tabular}{cc|ccccc}
			\toprule
			\multirow{2}{*}{Method} & \multirow{2}{*}{Year} & \multicolumn{5}{c}{Dataset} \\
			\cmidrule(r){3-7}
			&  & UCF-Crime(AUC) & XD-Violence(AP) & Ped2(AUC) & Avenue(AUC) & SHTech(AUC) \\
			\midrule
			
			Unmasking~\cite{208_tudor2017unmasking} & 2017 & / & / & 82.2\% & 80.6\% & / \\
			DAW~\cite{210_wang2018detecting} & 2018 & / & / & 96.4\% & 85.3\% & / \\
			
			MC2ST~\cite{209_liu2018classifier} & 2018 & / & / & 87.5\% & 84.4\% & / \\
			
			STDOR~\cite{211_pang2020self} & 2020 & / & / & 83.2\% & / & / \\
			CIL~\cite{213_lin2022causal} & 2022 & / & / & 99.4\% & 90.3\% & / \\
			LBR-SPR~\cite{214_yu2022deep} & 2022 & / & / & 97.2\% & 92.8\% & 72.6\% \\
			
			C2FPL~\cite{216_al2024coarse} & 2024 & 80.7\% & 80.\% & / & / & / \\
			CLAP~\cite{al2024collaborative} & 2024 & 83.2\% & 85.7\% & / & / & / \\
			
			CKNN~\cite{cknn} & 2024 & / & / & / & 94.1\% & 89.0\% \\
			InterUVAD~\cite{nie2024interleaving} & 2024 & / & / & / & / & 88.2\% \\
			
			\bottomrule
		\end{tabular}
	}
\end{table}

\section{Training-free VAD}
\label{sec:sec:tr}

In the field of VAD, severe imbalance between positive and negative samples, as well as difficulties in data annotation, have long been significant factors limiting model performance. With the rapid development of Large Language Models (LLMs) and Multimodal Large Language Models (MLLMs), researchers have discovered that the vast amount of pre-trained knowledge embedded in these models can be leveraged for anomaly detection without the need for additional model training. In this context, training-free VAD (TVAD) has emerged. The core idea of TVAD is to utilize the semantic interaction capabilities and multimodal understanding abilities of LLMs and MLLMs to first summarize videos into specific captions, and then analyze these captions to detect anomalies. It is worth noting that, although many works employing such models claim to belong to semi-supervised or weakly supervised VAD, they merely use the same types of data annotations as other VAD paradigms, without any actual model training. Therefore, we also categorize these methods within the scope of TVAD.

\subsection{Paradigm}

Since model training is not required, the core idea of TVAD lies in how to exploit the potential of MLLMs to better summarize or analyze video content. Specifically, MLLMs first perform video captioning or video QA tasks on the input video sequence or the entire video. Subsequently, additional LLMs or MLLMs further refine and clean the generated text, and this process may be repeated multiple times. Finally, based on the dataset or additional human-provided prompts, the LLM or MLLM analyzes the processed video content and outputs the probability of anomalies.

LAVAD~\cite{lavad} was the first to propose a TVAD paradigm that employs pre-trained LLMs and VLMs for temporal aggregation, offering explanations for anomalies while maintaining competitive performance. SUVAD~\cite{suvad} builds upon LAVAD, introducing a coarse-to-fine anomaly analysis and smoothing module to mitigate the hallucination problem. In addition, SUVAD leverages the video analysis capabilities inherent to MLLMs, enabling the model to autonomously determine the distinction between normal and anomalous events within a dataset. AnomalyRuler~\cite{followtherules_yang2024follow} provides a rule-based reasoning approach that demonstrates strong performance in static scenes. AnyAnomaly introduces customizable video anomaly detection (C-VAD) techniques and models, treating user-defined texts as anomalous events and effectively implementing anomaly detection through context-aware visual question answering.

VADSK~\cite{vadsk} designs a two-stage process of deduction and inference, detecting anomalies in surveillance videos through keyword identification. VERA~\cite{vera_ye2025vera} focuses on prompt engineering, proposing a language-based learning framework in which prompts are treated as learnable parameters and are continuously optimized; several targeted prompts can significantly improve anomaly detection performance. MCANet~\cite{mcanet} proposes a Multi-modal Caption-Aware Network, which aggregates existing VLMs, ALMs, and LLMs to dynamically generate and analyze textual descriptions of video frames for anomaly detection.

\subsection{Video Input}

The core idea of TVAD lies in leveraging the pre-trained knowledge of MLLMs to comprehensively understand and analyze video content. However, in most scenarios—whether in datasets or real-world applications—anomalous events only account for a small fraction of the overall temporal duration. Furthermore, due to computational constraints, current mainstream MLLMs are unable to perform detailed analysis on every frame of an entire video and instead rely on sampling strategies. More importantly, anomaly-related content constitutes only a minimal portion of the pre-training data for MLLMs. As a result, directly inputting the whole video into an MLLM often fails to yield satisfactory results. Currently, almost all MLLM-based TVAD methods adopt a strategy of segmenting videos prior to detailed analysis and reasoning~\textbf{(Segment/Frame-wise Video Feed)}. Although this approach does not offer advantages in inference speed, its training-free nature and strong generalization across scenarios compensate for these shortcomings.

Furthermore, MLLMs capable of performing Video QA or Video Captioning are predominantly pre-trained on RGB video streams. To better accommodate the inherent characteristics of these models and to fully exploit their capabilities, the most common approach at present is to segment the original \textbf{RGB} video and input these segments into the model.

LAVAD~\cite{lavad} segments videos into non-overlapping, fixed-length clips and then uses a captioning model to perform detailed analysis on each frame. AnomalyRuler~\cite{followtherules_yang2024follow}, AnyAnomaly~\cite{anyanomaly}, VADSK~\cite{vadsk}, and VERA all adopt this approach. SUVAD~\cite{suvad} further extends this by first conducting segment-level analysis to identify regions with a high probability of anomalies, followed by detailed frame-level detection within those regions. MCANet~\cite{mcanet} simultaneously inputs video segments, individual frames, and audio for multi-stream analysis.

\subsection{Model Architecture}

\subsubsection{LLM/MLLM}

As mentioned above, TVAD relies on the pre-trained knowledge of LLMs/MLLMs for video analysis. In this context, MLLMs with strong video content analysis capabilities and LLMs adept at uncovering hidden anomalous clues are favored by researchers. LAVAD~\cite{lavad} employs BLIP-2~\cite{blip2} and Llama-2-13b-chat~\cite{llama2} as the models for video understanding and content analysis, respectively. SUVAD~\cite{suvad} utilizes GLM-4V~\cite{glm} and Llama-3-7b~\cite{llama3} for these purposes. VERA ~\cite{vera_ye2025vera} evaluates the anomaly video analysis capabilities of InternVL2-8B~\cite{internvl}, InternVL2-40B~\cite{internvl}, and Qwen2-VL-7B~\cite{qwen2vl}. In addition, many other mainstream MLLMs, such as Video-ChatGPT~\cite{videochatgpt}, VTimeLLM~\cite{vtimellm}, Qwen2.5-VL~\cite{qwen2.5vl}, and TimeChat~\cite{timechat}, are widely applied in VAD.

\subsubsection{VLM}

Although LLMs/MLLMs possess strong reasoning abilities and robust generalization across scenarios, their inherent hallucination problem can severely impact anomaly detection performance. To mitigate the confusion caused by hallucinations, many approaches combine VLMs with LLMs/MLLMs. LAVAD~\cite{lavad} employs CLIP~\cite{clip} for segment-level and frame-level label cleaning to obtain more accurate content descriptions. AnyAnomaly~\cite{anyanomaly} uses CLIP to guide the model’s attention to the main content of the scene, thereby minimizing irrelevant background interference.

\subsection{Model Optimization}

Since it is not possible to train the model in TVAD tasks, and it is also difficult to optimize the model within the visual space, efficiently leveraging the instruction-following capabilities of LLMs/MLLMs is crucial for improving detection performance \textbf{(Semantic Space Optimization)}. Currently, there are two mainstream approaches. Prompt engineering focuses on optimizing the input queries to guide the model’s attention to key content within the video. Prior knowledge bases, on the other hand, artificially constrain the scope of analysis through various methods, converting open-set anomaly detection into closed-set anomaly detection to achieve better performance.

\subsubsection{Prompt Engineering}

Prompt engineering can play a significant role in nearly any domain related to LLMs/MLLMs. In VAD, the focus of prompt engineering is on leveraging the model’s instruction-following abilities to predefine the types of anomalies to be analyzed, the expected output format, and the irrelevant information to be ignored. VERA~\cite{vera_ye2025vera} utilizes a continuously updated set of questions to determine which types of anomalies should be prioritized. Works such as AnomalyRule~\cite{followtherules_yang2024follow}, AnyAnomaly~\cite{anyanomaly}, and LAVAD~\cite{lavad} further narrow the detection scope by predefining the anomalies to be detected or delineating normal events in advance, thereby enabling more targeted anomaly detection.

\subsubsection{Prior Knowledge Base}

Prior knowledge bases help LLMs and MLLMs better define the scope of anomalies and identify information that should be ignored. Generally, these knowledge bases are either manually specified or extracted from datasets using other models, and then incorporated into prompts as contextual information. For instance, SUVAD~\cite{suvad} and AnomalyRuler~\cite{followtherules_yang2024follow} use MLLMs to generate lists of normal events from the training set, providing a clear reference for what should be considered non-anomalous. VERA~\cite{vera_ye2025vera}, on the other hand, continually updates its question set based on training videos, ensuring the prompts remain precise and focused on relevant anomalous content.

\subsection{Performance Comparison and Paradigm Example}

\begin{figure}[h]
	\centering
	\includegraphics[width=\linewidth]{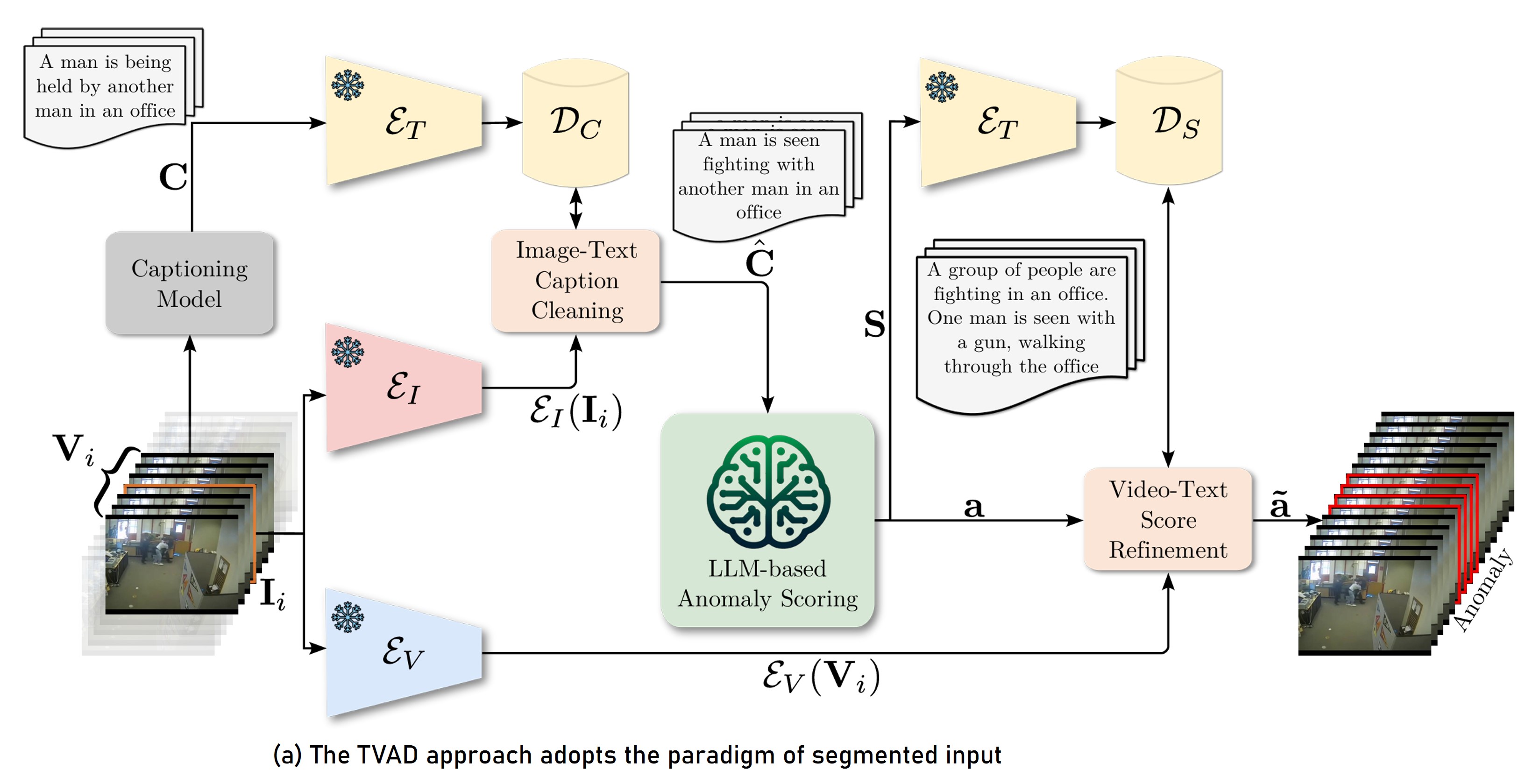}
	\caption{Fig (a) illustrates the TVAD method adopts the paradigm of segmented input~(LAVAD~\cite{lavad}).}
	\label{fig:tvad}
\end{figure}

We provide a performance comparison of existing methods in Table~\ref{table:tvad}, and present the classic paradigm of TVAD in Fig.~\ref{fig:tvad}. Due to space limitations, we did not include the performance comparison of all methods in the table. Instead, we selectively present the most advanced methods and highly cited classic methods.

\begin{table}
	\caption{Comparison of the performance of existing TVAD methods}
	\label{table:tvad}
	
	\resizebox{\linewidth}{!}{
		\begin{tabular}{cc|ccccc}
			\toprule
			\multirow{2}{*}{Method} & \multirow{2}{*}{Year} & \multicolumn{5}{c}{Dataset} \\
			\cmidrule(r){3-7}
			&  & UCF-Crime(AUC) & XD-Violence(AP) & Ped2(AUC) & Avenue(AUC) & SHTech(AUC) \\
			\midrule
			LAVAD~\cite{lavad} &2024 &80.28\% &62.01\% &/ &/ &/\\
			Anomaly Ruler~\cite{followtherules_yang2024follow} &2024 &/ &/ &97.40\% &81.60\% &83.50\%\\
			SUVAD~\cite{suvad} &2025 &83.90\% &70.10\% &96.80\% &89.30\% &80.20\%\\
			VADSK~\cite{vadsk} &2025 &/ &/ &86.50\% &74.20\% &75.30\%\\
			VERA~\cite{vera_ye2025vera} &2025 &86.55\% &88.26\%(AUC) &/ &/ &/\\
			
			\bottomrule
		\end{tabular}
	}
\end{table}

\section{Instruction Tuning VAD}
\label{sec:sec:in}

Compared to training-free VAD methods based on LLM/MLLM, constructing anomaly-related datasets and using them to perform instruction tuning on large models is also an important paradigm. Instruction tuning enables large models to transfer their general knowledge to specific anomaly detection tasks, thereby enhancing the model's understanding and discrimination of abnormal events. In recent years, with the continuous enrichment of open-source video anomaly datasets and advancements in multimodal annotation techniques, more and more researchers are exploring how to instruction-tune large models using high-quality abnormal videos and their language descriptions. Moreover, instruction tuning can not only directly update the parameters of the large model itself but also achieve efficient adaptation by freezing the large model and fine-tuning downstream modules. It is worth noting that model architectures under the instruction tuning paradigm are becoming increasingly diverse, including multimodal fusion architectures based on VLMs and LLMs, as well as hybrid models that combine GNNs, Transformers, or RAG to enhance reasoning capabilities.

With its flexible data-driven characteristics and strong knowledge transfer capabilities, instruction-tuned VAD is becoming an important development direction in the field of video anomaly detection, driving continuous improvements in accuracy, efficiency, and interpretability of VAD models.

\subsection{Paradigm}

\subsubsection{Fine-tuning LLM/MLLM}

In instruction tuning VAD~(ITVAD) tasks, directly fine-tuning LLMs/MLLMs is the most natural and relatively easy-to-implement approach. Building upon the general visual and language knowledge already present in such models, these methods further enable the model to learn the spatiotemporal features, semantic descriptions, and reasoning abilities related to abnormal events through the construction or use of anomaly-related datasets and task instructions.

The specific process typically includes the following key steps:
\begin{itemize}
	\item Dataset Construction and Instruction Design: Researchers first organize or create high-quality anomaly video datasets with textual annotations. These datasets not only contain video clips but are also accompanied by rich event descriptions, question-answer pairs, or hierarchical labels, providing large models with multi-perspective and multi-granularity learning signals.
	\item Fine-tuning Strategy Selection: Various strategies can be used, such as full-parameter fine-tuning, parameter-efficient fine-tuning (e.g., Adapter, LoRA), or instruction tuning, allowing flexible choices based on computing resources and task requirements.
	\item Multimodal Fusion and Spatiotemporal Modeling: By combining multimodal inputs such as video frames, optical flow, and semantic text, researchers design appropriate task instructions and output formats to enhance the model’s spatiotemporal understanding and semantic interpretability of abnormal events.
\end{itemize}

HAWK~\cite{hawk_tang2024hawk} achieved SOTA performance on open-world anomaly detection and question-answering explanation tasks by fine-tuning on large-scale anomaly videos and diverse QA pairs, and by constructing auxiliary consistency loss in motion and video space. UCVL~\cite{ucvl} improved the understanding of complex abnormal events by fine-tuning models like LLaVA-OneVision~\cite{llava-onevision}. AssistPDA~\cite{assistpda} proposed a spatiotemporal relation distillation module to transfer the long-term spatiotemporal modeling capability of VLM from offline settings to real-time scenarios, and built the first online VAD dataset, VAPDA-127K. By fine-tuning on this dataset, AssistPDA unified video anomaly prediction, detection, and analysis within a single framework. Holmes-VAU~\cite{holmes-vau_zhang2025holmes} constructed HIVAU-70k and combined the traditional weakly-supervised paradigm with MLLM fine-tuning, significantly improving anomaly detection performance.

\subsubsection{Frozen LLM/MLLM \& Efficient Module Tuning}

In reality, current image-centric MLLMs are already highly developed and can handle tasks such as image understanding and image QA at a high level. However, due to computational overhead constraints, video large models often need to compress the input video content through sampling or other means. This sampling process can lead to a significant loss of anomaly cues and cause the model to misinterpret event content.

Based on this, another important paradigm for instruction-tuned VAD tasks is to freeze the parameters of large models and use them as fixed feature extractors or knowledge sources, while only fine-tuning upstream or downstream modules. This approach balances the powerful abstract representation capabilities of large models with the efficient adaptation needs of upstream and downstream tasks, making it especially suitable for scenarios with limited computational resources, real-time inference, or edge deployment.

Specific implementation methods for this paradigm include:
\begin{itemize}
	\item Specialized Video Sampler Learning: Independently train a video sampler to extract frames with a high probability of anomaly occurrence and feed them into subsequent MLLM processing, providing more anomaly cues and improving detection performance.
	\item Efficient Fine-tuning of Downstream Modules: Pass the features or knowledge graphs output by the large models to lightweight downstream modules (such as GNNs, Transformers, retrieval networks, etc.), and only train or fine-tune these modules to adapt to specific anomaly detection tasks.
	\item Dynamic Adaptation and Incremental Learning: After freezing the large model, the downstream modules can quickly adapt to new anomaly types or behavioral patterns through online learning, graph updates, or module augmentation, thereby enhancing the model’s flexibility and robustness.
\end{itemize}

SlowFastVAD~\cite{slowfastvad} uses a lightweight fast detector in collaboration with a RAG-enhanced VLM. The VLM is only invoked for fine-grained analysis and reasoning when a suspicious segment is detected, greatly improving inference efficiency and system scalability. MissionGNN~\cite{missiongnn} dynamically updates the knowledge graph structure on edge devices, achieving continual adaptation and efficient inference. Vad-llama~\cite{vad-llama} adopts a three-stage training process, combining long-term context modules, video anomaly detectors, anomaly prediction variables, and projection layers with Llama, achieving strong performance on benchmarks like UCF-Crime and TAD. VLAVAD~\cite{vlavad} leverages the reasoning ability of LLMs and selective prompt adapters (SPA) to choose the semantic space and introduces a sequence state space module, significantly enhancing the interpretability of anomaly detection. CUVA~\cite{cuva_du2024uncovering} instruction-tunes a MIST~\cite{mist} selector to adaptively select video frames likely to contain anomalies and feeds them into the frozen MLLM, enriching the extracted anomaly cues and significantly improving the understanding of video details and the ability to detect anomalies in long videos.

\subsection{Video Input}

\subsubsection{Complete Video}

Compared to most traditional VAD methods that utilize segment-wise or frame-wise input, instruction-tuned VAD fully leverages the powerful capabilities of large models in temporal modeling and global semantic understanding, enabling holistic input and processing of complete long video segments. By directly performing anomaly comprehension on entire video sequences, this approach not only significantly alleviates the inference speed bottleneck commonly associated with large models but also yields notable improvements in anomaly event detection performance compared to training-free VAD methods. Instruction tuning not only enhances detection accuracy but also further strengthens the model’s global perception and interpretive abilities in complex anomalous scenarios.

\textbf{Uniform Sampling}.
Analogous to existing Visual Question Answering (VQA) tasks, uniformly sampling frames from an entire video segment without any additional explicit prompts represents the most fundamental and straightforward approach to reducing computational costs.

For instance, AssistPDA~\cite{assistpda} employs the Qwen2-VL~\cite{qwen2vl} visual encoder to process both segmented raw videos and consecutive video frames, aligning the extracted CLS tokens via the STRD module before feeding them into the LLM for inference. Similarly, HAWK~\cite{hawk_tang2024hawk} utilizes the EVA CLIP~\cite{eva_clip} encoder together with a Q-former to uniformly extract features from both video frames and optical flow. SlowFastVAD~\cite{slowfastvad} directly adopts the original sampling strategy of the VLM without further modification.

This approach incurs no additional overhead in module design or system complexity, making it highly straightforward from an engineering perspective. However, considering that anomalous events account for only a small proportion of the data in the field of video anomaly detection (VAD), uniform sampling often selects normal frames from within anomalous video segments, which may mislead the model and result in incorrect interpretations.

\textbf{Non-uniform Sampling}.
Since uniform sampling inevitably introduces a considerable amount of irrelevant interference, guiding the model to focus on segments with a higher likelihood of containing anomalies—rather than unrelated regions—has become a crucial factor in improving the performance of instruction-tuned VAD methods.

The adaptive sampling technique employed in Holmes-VAU~\cite{holmes-vau_zhang2025holmes} dynamically attends to regions of interest along the temporal axis. By prioritizing frames that are more likely to contain anomalies, this approach enables finer-grained detection and greater computational efficiency.
Similarly, CUVA~\cite{cuva_du2024uncovering} fine-tunes a MIST~\cite{mist} module to select tokens that are specifically relevant to anomalous events in the video.

Such adaptive strategies achieve an optimal balance and perform particularly well when anomalies are temporally sparse or highly context-dependent. While this approach substantially enhances the detection performance of the model, it also inevitably introduces additional computational overhead and increases training costs due to the integration of these guiding sampling modules.

\subsubsection{Segment/Frame-wise Video Feed}

Similar to training-free VAD approaches, some methods utilize dense video segmentation combined with instruction tuning, aiming to further enhance the model's capability to interpret anomalous events.

For instance, Vad-llama~\cite{vad-llama} divides the original video into equal-length segments, feeds them into the model, and employs a three-stage fine-tuning strategy, resulting in notable improvements in performance.

\subsection{Model Architecture}

As the absolute core of ITVAD, the general capabilities of MLLMs significantly influence the detection performance of the fine-tuned models. Currently, the mainstream approaches employ representative multimodal large models such as LLava-1.5, Qwen2-VL, CogVLM, and InternVL2, which demonstrate excellent alignment and understanding of both visual and linguistic modalities, thereby providing a solid foundation for downstream tasks.

For example, CUVA~\cite{cuva_du2024uncovering} adopts VideoChatGPT~\cite{videochatgpt} as a frozen backbone model and performs fine-tuning exclusively on the additional MIST module~\cite{mist}, fully leveraging the powerful representational capacity of the pretrained model. Holmes-VAU~\cite{holmes-vau_zhang2025holmes} utilizes InternVL2~\cite{internvl} as the primary MLLM and incorporates LoRA-based fine-tuning to better adapt the model to anomaly understanding tasks across multiple temporal granularities. AssistPDA~\cite{assistpda} employs Qwen2-VL~\cite{qwen2vl} as the visual encoder and extends its offline temporal modeling capability to online inference frameworks through the Spatio-Temporal Relation Distillation (STRD) module, thereby enhancing the model’s real-time inference capabilities. In contrast, HAWK~\cite{hawk_tang2024hawk} does not directly use an MLLM; instead, it adopts a lightweight GNN architecture, integrating LLMs and vision-language models such as EVA CLIP~\cite{eva_clip} to achieve efficient and dynamic adaptation to anomalous events in open scenarios, thus improving the model’s flexibility and generalization in practical applications.

\subsection{Model Optimization}

\subsubsection{Visual Space Optimization}

In the context of visual domains, large models leverage their powerful multimodal information processing capabilities to enable more efficient and accurate applications of pseudo-anomaly generation, memory bank design, and spatio-temporal modeling. The dynamic generation of pseudo-anomalies, when combined with the visual generalization ability of large models, allows training processes to rapidly adapt to complex and dynamic scenarios. Moreover, the integration of memory mechanisms with spatio-temporal modeling modules enables large models to more precisely focus on key regions and capture subtle variations within visual patterns, thereby improving the accuracy of anomaly detection.

\textbf{Pseudo Anomalies}.
Pseudo-anomaly generation is a key technique in SVAD, as it enables the simulation of abnormal distributions by synthesizing artificial anomaly samples to enhance training effectiveness. This process strengthens the model’s ability to detect previously unseen anomalies. HAWK~\cite{hawk_tang2024hawk} adopts and extends this concept by dynamically adjusting and perturbing nodes within its knowledge graph to generate diverse pseudo-anomalous data. Such an approach not only enriches the distribution of training samples, but also allows the model to better adapt to complex and evolving dynamic environments during training. As a result, the model’s generalization capability and robustness in real-world scenarios are significantly improved.

\textbf{Memory Bank}.
The memory bank plays a crucial role in capturing the distribution of normal behaviors. By efficiently storing and retrieving the feature representations of normal samples, the memory bank provides strong support for anomaly detection tasks. Holmes-VAU~\cite{holmes-vau_zhang2025holmes} leverages the memory module to store a large number of normal sample features and, during inference, compares the features of test samples with those stored in the memory bank. By maximizing the distance between the features of normal and abnormal samples, this approach effectively enhances the model’s ability to distinguish anomalies and improves the accuracy of anomaly detection.

\textbf{Spatio-Temporal Modeling}.
Spatio-temporal modeling is critical for enhancing the model’s ability to focus on key regions within the input data. By jointly modeling spatial and temporal information, the model can more comprehensively understand dynamic changes and salient events in video sequences. AssistPDA~\cite{assistpda} incorporates the Spatio-Temporal Relation Distillation (STRD) module to effectively integrate visual inputs with temporal information, thereby enhancing the model’s awareness of contextual cues before and after anomalous events. This enables the model to capture subtle spatial-temporal relationships associated with anomalies.

\subsubsection{Semantic Space Optimization}

In semantic space optimization, the LLMs' robust language comprehension and knowledge integration capabilities are particularly critical. Through prompt engineering, LLMs can extract high-quality anomaly analysis results from multi-level linguistic cues, while also responding more effectively to complex user instructions. The dynamic integration of prior knowledge bases further enhances the model's reasoning ability, enabling strong adaptability when encountering previously unseen anomaly patterns. These optimization strategies fully leverage large models’ strengths in semantic reasoning and knowledge transfer, thus providing robust support for anomaly detection in complex scenarios.

\textbf{Prompt Engineering}.
Similar to TVAD, prompt engineering also plays an essential role in ITVAD. Carefully designed and optimized prompts can effectively guide the model to understand task requirements and input content, resulting in more accurate detection and analysis outcomes. For example, Holmes-VAU~\cite{holmes-vau_zhang2025holmes} employs a hierarchical prompt design, including prompts for Caption, Judgment, and Analysis, which progressively guide the model to generate high-quality anomaly analysis results. AssistPDA~\cite{assistpda} enhances the model’s ability to respond to complex user queries by optimizing natural language prompts, thereby improving its capability to identify and interpret anomalous events. CUVA~\cite{cuva_du2024uncovering} introduces a multi-turn question optimization mechanism, using a sequence of targeted questions to help the model focus on abnormal phenomena in videos, thus enhancing both the effectiveness of anomaly detection and the model’s interactivity.

\textbf{Prior Knowledge Base}.
The introduction of knowledge bases significantly enhances the model’s ability to understand complex scenarios. By integrating structured knowledge, the model can dynamically update abnormal patterns in the environment and, combined with commonsense reasoning, efficiently detect previously unseen anomalies. HAWK~\cite{hawk_tang2024hawk} leverages knowledge graphs for dynamic modeling and management of abnormal events, enabling rapid adaptation to newly emerging anomalies. Meanwhile, its lightweight design ensures efficient inference and deployment on edge computing devices.

\subsection{Performance Comparison and Paradigm Example}

\begin{figure}[h]
	\centering
	\includegraphics[width=\linewidth]{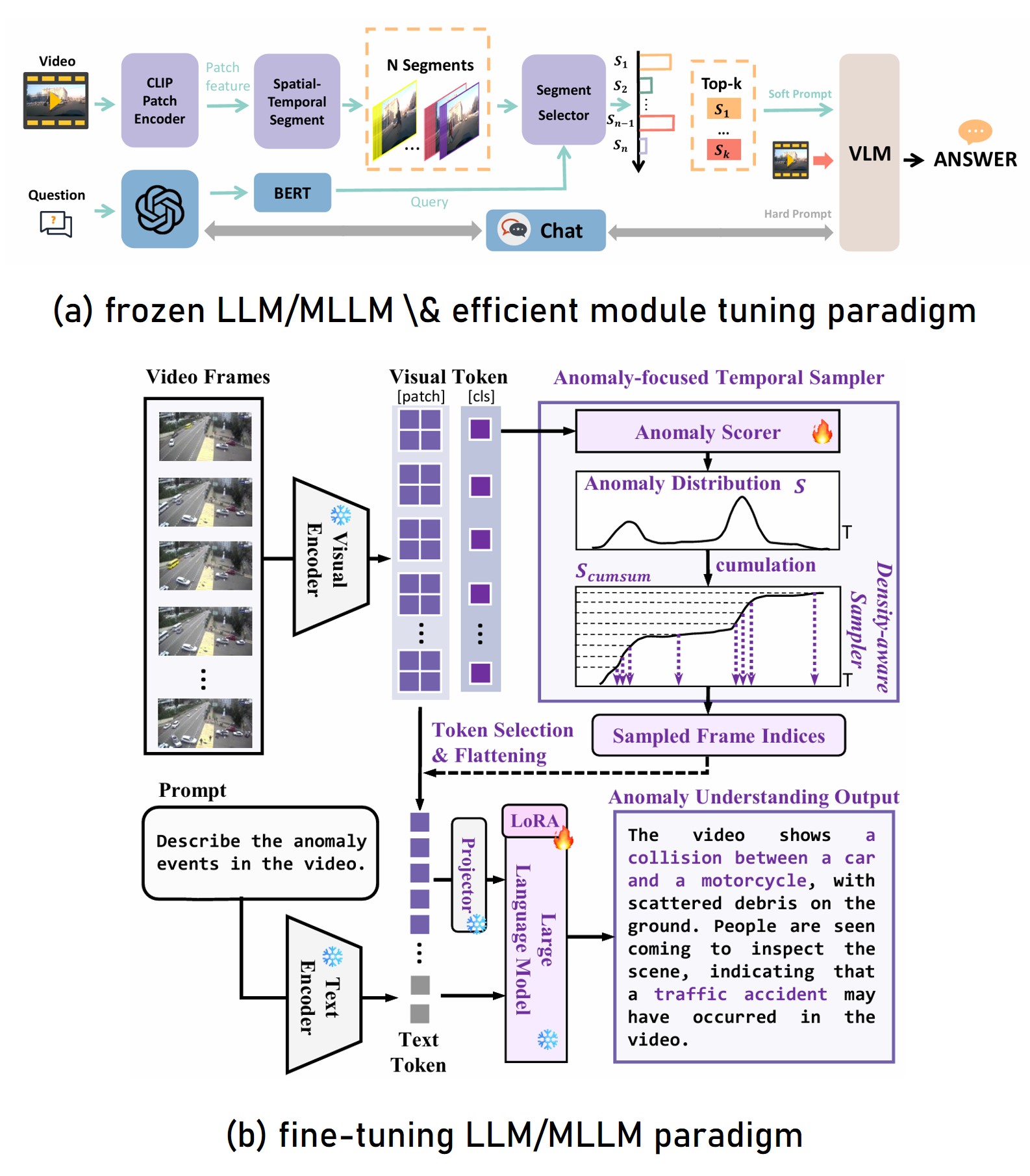}
	\caption{Fig (a) illustrates the frozen LLM/MLLM \& efficient module tuning paradigm~(CUVA~\cite{cuva_du2024uncovering}), while Fig (b) shows the fine-tuning LLM/MLLM paradigm (Holmes-VAU~\cite{holmes-vau_zhang2025holmes}).}
	\label{fig:itvad}
\end{figure}

We provide a performance comparison of existing methods in Table~\ref{table:itvad}, and present the classic paradigm of ITVAD in Fig.~\ref{fig:itvad}. Due to space limitations, we did not include the performance comparison of all methods in the table. Instead, we selectively present the most advanced methods and highly cited classic methods.

\begin{table}
	\caption{Comparison of the performance of existing ITVAD methods}
	\label{table:itvad}
	
	\resizebox{\linewidth}{!}{
		\begin{tabular}{cc|ccc}
			\toprule
			Method & Year & \multicolumn{3}{c}{Dataset \& Performance} \\

			\midrule
			
			VADLlama~\cite{vad-llama} &2024 &TAD(AUC) 88.13\% &UCF-Crime(AUC) 91.77\% &\\
			CUVA~\cite{cuva_du2024uncovering} &2024 &CUVA(MMEval) 79.65 58.92 50.64 & &\\
			HAWK~\cite{hawk_tang2024hawk} &2024 &HAWK(GPT-Guided) 0.283 0.320 0.218 & &\\
			VLAVAD~\cite{vlavad} &2024 &Ped2(AUC) 99.00\% &Avenue(AUC) 87.6\% &SHTech(AUC) 87.2\% \\
			
			Holmes-VAU~\cite{holmes-vau_zhang2025holmes} &2025 &HIVAU-70k(BLEU) 0.916 0.804 0.566 &UCF-Crime(AUC) 88.96\% &XD-Violence(AP) 87.68\\
			UCVL~\cite{ucvl} &2025 &UCVL(GPT-4o) 63.8 & &\\
			
			SlowFastVAD~\cite{slowfastvad} &2025 &Ped2(AUC) 99.10\% &Avenue(AUC) 89.6\% &SHTech(AUC) 85.0\% \\
			MissionGNN~\cite{missiongnn} &2025 &UCF-Crime Incomplete(AUC) 91.00\% & &\\

			\bottomrule
		\end{tabular}
	}
\end{table}

\section{Open-set VAD}
\label{sec:sec:ops}

Deploying well-trained supervised models in real-world scenarios to detect previously unseen anomalies is a critical step for the widespread application of VAD in practice. In real-world environments, it is impossible to anticipate all possible anomalous events in advance; thus, open-set VAD (OSVAD) has emerged to address this challenge. Unlike traditional closed-set VAD, where the types of anomalies are known and clearly defined, OSVAD must handle unforeseen and unknown anomalies. Compared to semi-supervised and weakly supervised VAD, research on OSVAD remains relatively limited. Moreover, with the rapid development of VLMs and LLMs in recent years, OSVAD is gradually being replaced by training-free VAD and instruction-tuned VAD. Due to the scarcity of research in this area, we provide a detailed description of the few existing methods here, rather than categorizing them according to the framework proposed earlier. Generally, OSVAD can be divided into two main types: open-set VAD and few-shot VAD.

\subsection{Open-set VAD}

OSVAD focuses on discovering anomalous events that were not observed during training, thereby overcoming the limitations of traditional VAD, which only targets known types of anomalies. In real-world scenarios, the diversity and unpredictability of anomaly types make OSVAD particularly significant for practical applications. Representative work in this area includes MLEP~\cite{219_liu2019margin}, which was the first to propose the open-set supervised VAD paradigm. MLEP aims to effectively distinguish between normal and anomalous samples by learning appropriate margins in the feature space, even when only a very limited number of anomalous samples are available. Subsequently, the introduction of the UBnormal benchmark~\cite{ubnormal} provided a unified platform for the evaluation and comparison of OSVAD methods, promoting the standardization and systematic development of this field. In addition, Zhu et al.~\cite{221_zhu2022towards} proposed using normalized flow models to generate pseudo-anomalous features, which significantly improves the generalization and detection capabilities for previously unseen anomaly types. Some studies~\cite{222_ding2022catching,223_zhu2024anomaly}, although focusing on open-set anomaly detection at the image level, also offer valuable insights for research in video settings. Overall, OSVAD has notably enhanced the ability to detect unknown anomalies through innovative feature learning, benchmark construction, and generalization strategies, thus laying a solid foundation for the application of VAD in complex and open environments.

\subsection{Few-shot VAD}

The primary objective of few-shot VAD~(FSVAD) is to achieve anomaly detection in a target scenario when only a very limited number of frames containing previously unseen anomalous events are provided. In contrast to OSVAD, FSVAD assumes that a small number of real anomalous samples from the target scenario are available during the testing phase. This task was first introduced by Lu et al.~\cite{224_lu2020few}, who adopted a meta-learning model to enable rapid adaptation to new scenarios, requiring fine-tuning with a small set of samples. To enhance practicality and avoid additional fine-tuning before deployment, subsequent works such as those by Hu et al.~\cite{225_hu2021adaptive} and Huang et al.~\cite{226_huang2022boosting} proposed adaptive methods based on metric learning and variational networks, which leverage a small number of normal samples and enable inference in new scenarios without model fine-tuning. Furthermore, Aich et al.~\cite{227_aich2023cross} proposed the zxVAD framework, which achieves anomaly detection across domains in an unsupervised manner. This approach innovatively introduces untrained CNNs to generate pseudo-anomalous samples, eliminating the need for target domain adaptation. Overall, FSVAD emphasizes improving model generalization and transfer capabilities under conditions of extremely limited or even no labeled data, through approaches such as meta-learning or feature adaptation, thus providing effective solutions for anomaly detection in data-scarce real-world scenarios.

\subsection{Performance Comparison and Paradigm Example}

\begin{figure}[h]
	\centering
	\includegraphics[width=\linewidth]{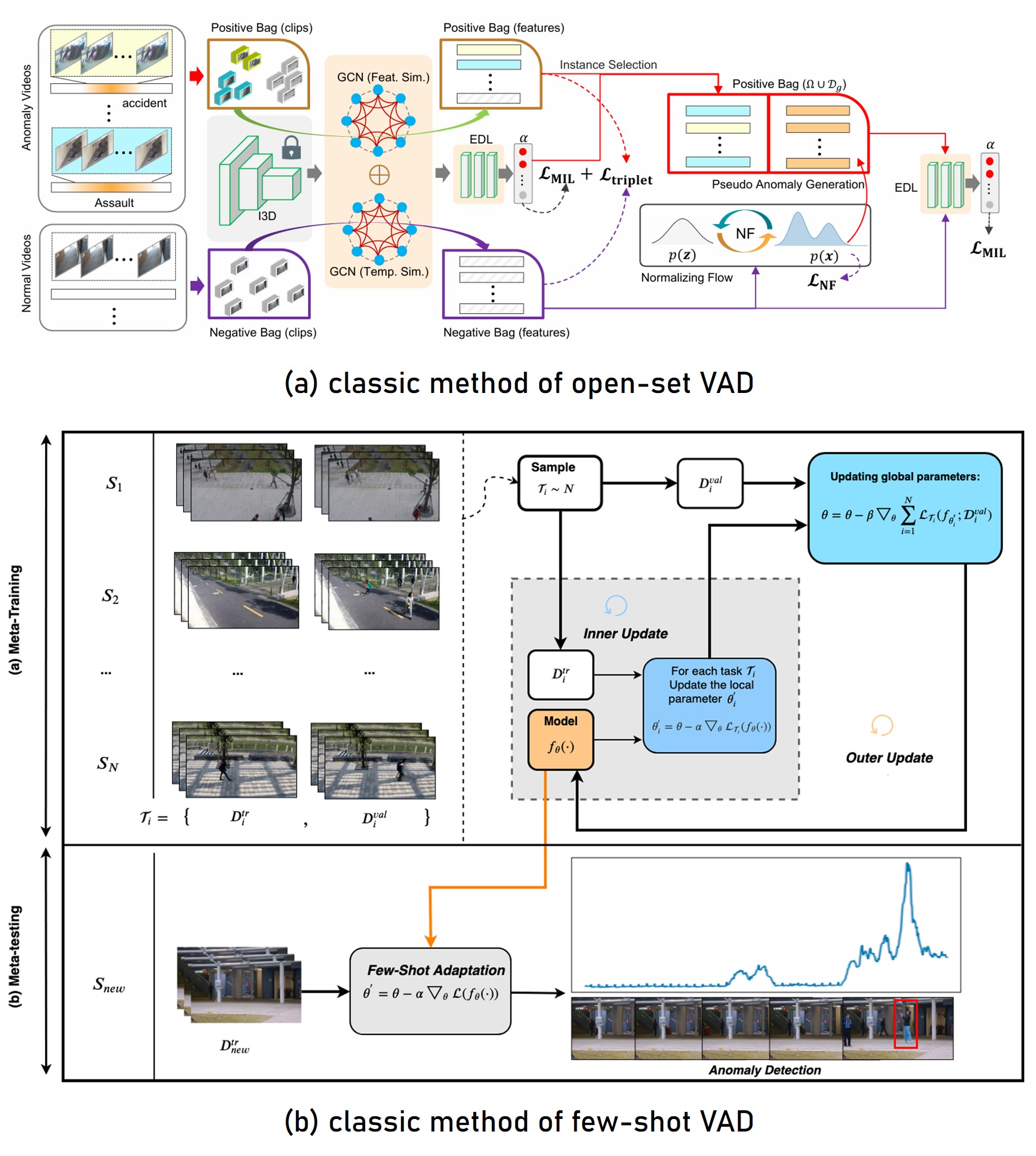}
	\caption{Fig (a) illustrates the classic method of OSVAD~(Zhu et al.~\cite{221_zhu2022towards}), while Fig (b) shows the classic method of few-shot VAD (Hu et al.~\cite{225_hu2021adaptive}).}
	\label{fig:osvad}
\end{figure}

We provide a performance comparison of existing methods in Table~\ref{table:osvad}, and present the classic paradigm of OSVAD in Fig.~\ref{fig:osvad}. Due to space limitations, we did not include the performance comparison of all methods in the table. Instead, we selectively present the most advanced methods and highly cited classic methods.

\begin{table}
	\caption{Comparison of the performance of existing OSVAD methods}
	\label{table:osvad}
	
	\resizebox{\linewidth}{!}{
		\begin{tabular}{cc|ccccc}
			\toprule
			\multirow{2}{*}{Method} & \multirow{2}{*}{Year} & \multicolumn{5}{c}{Dataset} \\
			\cmidrule(r){3-7}
			&  & UCF-Crime(AUC) & XD-Violence(AP) & Ped2(AUC) & Avenue(AUC) & SHTech(AUC) \\
			\midrule
			
			MLEP~\cite{219_liu2019margin} & 2019 &/ &/ &/ &92.80\% &76.80\% \\
			\multirow{2}{*}{Zhu et al.~\cite{221_zhu2022towards}} & \multirow{2}{*}{2021} &80.14\%
			 &69.61\% &/ &/ &/ \\
			& &(9 anomaly seen) &(4 anomaly seen) & & & \\
			
			\midrule
			
			\multirow{2}{*}{Hu et al.~\cite{225_hu2021adaptive}} & \multirow{2}{*}{2021} &/ &/ &96.20\% &85.80\% &77.9\% \\
			& & & &(13 sence seen) &(13 sence seen) &(13 sence seen) \\
			\multirow{2}{*}{Huang et al.~\cite{226_huang2022boosting}} & \multirow{2}{*}{2022} &/ &/ &95.12\% &82.62\% &/ \\
			& & & &(10 shot) &(10 shot) & \\
			Aich et al.~\cite{227_aich2023cross} & 2023 &/ &/ &96.95\% &/ &71.60\% \\
			
			\bottomrule
		\end{tabular}
	}
\end{table}

\section{Open-vocabulary VAD}
\label{sec:sec:opv}

\subsection{Methods}

In real-world anomaly detection tasks, traditional methods typically rely on predefined anomaly categories or specific event labels, which fall under the paradigm of closed vocabulary. However, in practical applications, the types of anomalies are diverse and often unpredictable, making it difficult for a limited set of labels to cover all potential anomalous scenarios. To address this challenge, open vocabulary VAD~(OVVAD) has emerged. OVVAD enables models to represent anomalous events using open-ended natural language descriptions, rather than being restricted to the labels or categories seen during training.

Unlike OSVAD, OVVAD focuses on the “semantic openness” of anomalous events, meaning that the detection model can understand and recognize a wide range of anomalies described freely in natural language. For example, users can query the system for anomalous events using novel descriptive phrases (such as “someone suddenly collapsing” or “appearance of an unauthorized vehicle”) without the need to predefine these specific anomaly labels.

It should be noted that OVVAD, as an extension of OSVAD, has considerable overlap in objectives and form with WVAD, instruction-tuned VAD, and TVAD. However, OVVAD is fundamentally distinct from these tasks in essential ways. Therefore, we categorize it as a separate class. Similar to the OSVAD, research on OVVAD remains limited. Thus, we provide a detailed description of the few existing methods in this area, rather than categorizing them according to the framework we proposed earlier.

Wu et al.~\cite{wu2024open} first proposed a systematic OVVAD framework, dividing the task into category-agnostic anomaly detection and category-aware anomaly recognition. They utilize large vision-language models (e.g., CLIP) to boost generalization and introduce a Semantic Knowledge Injection (SKI) module to enrich scene and action vocabularies using large language models. Additionally, a Novelty Anomaly Synthesis (NAS) module generates unseen anomalies with AIGC techniques to enhance anomaly representation.

Liu et al.~\cite{liu2025language} extended this by introducing LaGoVAD, a language-guided open-world anomaly detection method. Here, anomaly definitions are modeled as variable random variables, enabling users to specify anomaly criteria in natural language, which increases flexibility and interactivity. The method leverages vision-language alignment, contrastive learning, and the comprehensive PreVAD dataset.

Yang and Radke~\cite{yang2025detecting} approached the problem from a spatial context, using object behavior clustering to automatically discover activity patterns, reducing the need for manual labels and improving adaptation to diverse definitions of normal behavior.

Overall, OVVAD has driven the transition of video anomaly detection from “label-constrained” to “semantically open” paradigms. Notably, with the advancements in LLMs and MLLMs for visual and semantic understanding, the objectives of OVVAD are also continuously expanding and evolving.

\subsection{Performance Comparison and Paradigm Example}

\begin{figure}[h]
	\centering
	\includegraphics[width=\linewidth]{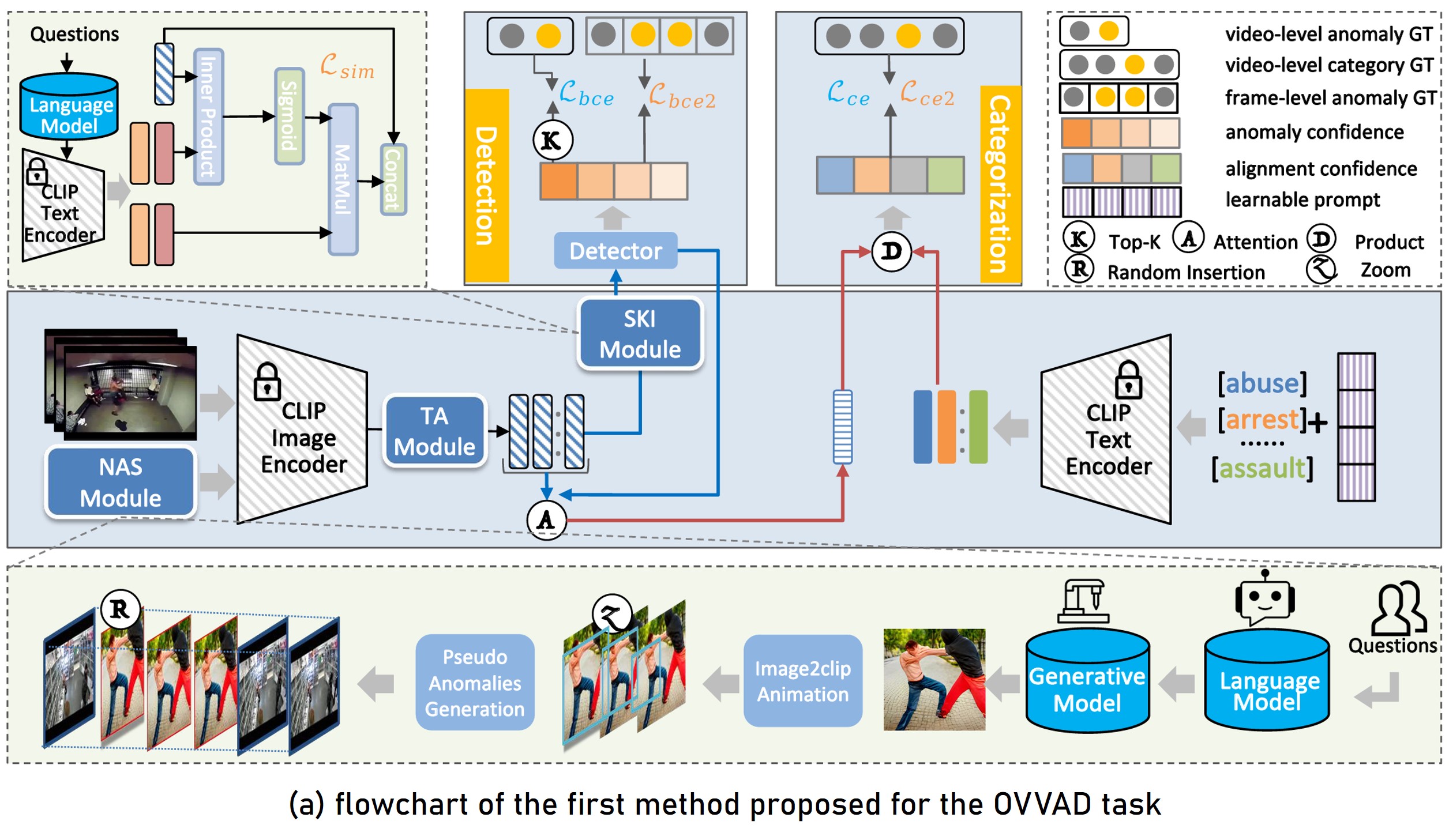}
	\caption{Fig (a) illustrates the flowchart of the first method proposed for the OVVAD task~(Wu et al.~\cite{wu2024open}).}
	\label{fig:ovvad}
\end{figure}

We provide a performance comparison of existing methods in Table~\ref{table:ovvad}, and present the classic paradigm of OVVAD in Fig.~\ref{fig:ovvad}. Due to space limitations, we did not include the performance comparison of all methods in the table. Instead, we selectively present the most advanced methods and highly cited classic methods.

\begin{table}
	\caption{Comparison of the performance of existing OVVAD methods}
	\label{table:ovvad}
	
	\resizebox{\linewidth}{!}{
		\begin{tabular}{cc|ccc}
			\toprule
			Method & Year & \multicolumn{3}{c}{Dataset \& Performance} \\
			\midrule
			
			Wu et al.~\cite{wu2024open} &2024 &UCF-Crime(AUC) 86.40\% &XD-Violence(AP) 66.53\% &UBnormal(AUC) 62.94\% \\
			LaGoVAD~\cite{liu2025language} &2025 &UCF-Crime(Acc) 51.72\% &XD-Violence(Acc) 78.13\% & \\
			Yang and Radke ~\cite{yang2025detecting} &2025 &Street Scene(AUC) 67.0\% &Ped2(AUC) 99.0\% &SHTech(AUC) 81.3\% \\

			\bottomrule
		\end{tabular}
	}
\end{table}

\section{Future Opportunities}

The emergence of large models as the mainstream paradigm in Video Anomaly Detection (VAD) is now an inevitable trend, bringing unprecedented capabilities while also introducing new challenges and research frontiers. Based on the current technological landscape and unsolved bottlenecks, we identify several promising future opportunities for VAD in the era of large models:

\subsection{Scaling to Larger, Multi-Modal, and More Diverse Datasets}

The future of VAD lies in exploiting ever-larger and more diverse datasets that encompass a wide range of modalities and real-world scenarios. While current benchmarks are limited in scale and diversity, next-generation datasets should integrate RGB, optical flow, skeleton, semantic maps, audio, and even 3D multi-view information. Such rich, multi-modal data will enable large models to learn robust and generalizable representations, facilitate cross-modal reasoning, and improve detection in complex, unconstrained environments. Furthermore, expanding data coverage to include rare anomaly types, long-tail events, and challenging contexts (e.g., occlusion, adverse weather, crowded scenes) will be critical for building truly reliable and practical VAD systems.

\subsection{Improving Explainability and Hallucination Suppression}

Interpretability is paramount for real-world adoption of VAD, especially in high-stakes applications such as surveillance and public safety. In the era of large models, explainability should evolve from merely identifying anomalies to providing comprehensive causal analysis—clarifying what happened, why it happened, and what the potential consequences are. This requires models to generate interpretable rationales, causal chains, and actionable insights. At the same time, hallucination remains a significant challenge for large language and vision-language models, potentially leading to unreliable or misleading anomaly explanations. Future research should focus on integrating retrieval-augmented generation (RAG), chain-of-thought reasoning, and knowledge-grounded modules to both enhance interpretability and effectively suppress hallucinations, thereby improving trustworthiness and accountability.

\subsection{Balancing Computational Efficiency and Accuracy}

Despite their superior generalization and reasoning abilities, large models are often associated with substantial computational costs, hindering their deployment in real-time or resource-constrained scenarios. Achieving an optimal balance between computational efficiency and detection accuracy is an urgent research priority. Promising directions include the development of training-free or lightweight methods that leverage large model priors for rapid inference, as well as instruction-tuning and parameter-efficient fine-tuning techniques (e.g., adapters, LoRA) to reduce resource consumption without sacrificing performance. Further exploration of model distillation, modular architectures, and dynamic inference strategies will also be crucial for making large model-based VAD practical at scale.

\subsection{Enhancing Generalization to Unseen Scenarios}

While large models have significantly improved the generalization ability of VAD, current systems still struggle to handle the full diversity of real-world environments and previously unseen anomaly types. Future research should investigate methods for open-set and open-vocabulary anomaly detection, domain adaptation, and continual or lifelong learning. Leveraging multi-modal pre-training, meta-learning, and prompt-based adaptation, models could dynamically adjust to new scenes, novel behaviors, and evolving anomaly definitions, ensuring robust performance across various operational contexts. Building benchmarks that systematically evaluate generalization to unseen domains and rare anomalies will further drive progress in this direction.

\subsection{Advancing the Intrinsic Anomaly Detection Capacity of Large Models}

In real-world applications, anomalous events are inherently rare, and labeled abnormal data is extremely scarce. This makes it imperative to enhance the intrinsic anomaly detection capabilities of large models, enabling them to recognize and reason about anomalies with minimal supervision. Future work should explore self-supervised and unsupervised learning paradigms that fully leverage the vast semantic and visual priors encoded in large models. Additionally, integrating out-of-distribution detection, uncertainty estimation, and memory-augmented mechanisms could further improve the model's ability to identify subtle or previously unseen anomalies. Research into multimodal alignment and knowledge transfer can also help large models generalize anomaly detection skills across tasks and domains.

\subsection{Other Emerging Directions}

Beyond the above, several other research avenues are gaining traction:
\begin{itemize}
	\item \textbf{Human-in-the-loop and Interactive VAD}: Incorporating human feedback for system adaptation, active learning, and improved interpretability.
	\item \textbf{Privacy-preserving and Federated Learning}: Enabling anomaly detection in distributed, privacy-sensitive environments.
	\item \textbf{Robustness, Fairness, and Security}: Addressing adversarial robustness, bias mitigation, and secure deployment of VAD systems.
\end{itemize}

\section{Conclusion}

This paper provides a systematic review of research progress in video anomaly detection (VAD) in the era of large language models (LLMs) and multi-modal large models (MLLMs). Unlike previous surveys that mainly focused on traditional deep learning or single paradigms, this work proposes a unified framework to help researchers build a knowledge system from a broader and more systematic perspective. Based on this framework, we systematically categorize existing VAD methods, covering a wide range of paradigms including semi-supervised, weakly supervised, unsupervised, open-set, open-vocabulary, training-free, and instruction-tuning approaches. For each category, we further analyze representative methods in terms of model architecture, input modality, optimization strategies, and ways of integrating large models. We also provide unified performance comparisons and summaries of mainstream methods, comprehensively revealing the strengths and limitations of each approach. Finally, this paper summarizes the key challenges facing the field and discusses future research directions for large model-driven VAD, with the aim of providing valuable references and inspiration for subsequent research and practical applications.

%
%
%
%
%
%
%
%
%
\bibliographystyle{IEEEtran}
\bibliography{paper6}

\end{document}